\documentclass[conference]{IEEEtran}
\IEEEoverridecommandlockouts

\usepackage[utf8]{inputenc} % Only if using PDFLaTeX
\usepackage[T1]{fontenc}
\usepackage{lmodern}

\usepackage{amsmath,amssymb,amsfonts,mathtools}

\usepackage{graphicx}

\usepackage[table,xcdraw]{xcolor}

\usepackage{tabularx}
\usepackage{multirow}
\usepackage{arydshln}
\usepackage{float}
\usepackage{placeins}
\usepackage{threeparttable}

\usepackage{algorithm}
\usepackage{algpseudocode}

\usepackage{url}
\usepackage[colorlinks,bookmarksopen,bookmarksnumbered,citecolor=red,urlcolor=red]{hyperref}

\usepackage[backend=bibtex,style=authoryear,uniquelist=false,mincitenames=1, maxcitenames=1,url=false]{biblatex}
\DeclareDelimFormat{nameyeardelim}{\addcomma\space}
\DeclareCiteCommand{\cite}
  {\boolfalse{citetracker}%
   \boolfalse{pagetracker}%
   \usebibmacro{prenote}}
  {\ifciteindex
     {\indexnames{labelname}}
     {}%
   \printtext[bibhyperref]{%
     \printnames{labelname}%
     \setunit{\nameyeardelim}%
     \printfield{year}}}
  {\multicitedelim}
  {\usebibmacro{postnote}}

\usepackage{titlesec}
\titlespacing*{\section}{0pt}{10pt}{3pt}
\titlespacing*{\subsubsection}{0pt}{6pt}{3pt}

\usepackage{authblk}
\usepackage{etoolbox}
\usepackage{pifont}
\usepackage{xpatch}

\DeclareMathOperator*{\argmax}{argmax}
\newcommand{\cmark}{\ding{51}}%
\newcommand{\xmark}{\ding{55}}%

\newenvironment{conditions*}
  {\par\vspace{\abovedisplayskip}\noindent
   \tabularx{\columnwidth}{>{$}l<{$} @{${}:{}$} >{\raggedright\arraybackslash}X}}
  {\endtabularx\par\vspace{\belowdisplayskip}}

\begin{document}

\title{TD-TOG Dataset: Benchmarking Zero-Shot and One-Shot Task-Oriented Grasping for Object Generalization}

\author[1]{Valerija Holomjova}
\author[2]{Jamie Grech}
\author[3]{Dewei Yi}
\author[4]{Bruno Yun}
\author[1]{Andrew Starkey}
\author[5]{Pascal Meißner}
\affil[1]{\footnotesize School of Engineering, University of Aberdeen, Scotland UK}
\affil[2]{\footnotesize Data Intelligence Hub, Bank of Valletta, Malta}
\affil[3]{\footnotesize Department of Computing Science, University of Aberdeen, Scotland UK}
\affil[4]{\footnotesize Universite Claude Bernard Lyon 1, CNRS, École Centrale de Lyon, INSA Lyon, Université Lumière Lyon 2, LIRIS, UMR5205, France}
\affil[5]{\footnotesize Center for Artificial Intelligence (CAIRO), Technical University of Applied Sciences Würzburg-Schweinfurt (THWS), Würzburg, Germany}

\maketitle

\begin{abstract}\footnotesize
Task-oriented grasping (TOG) is an essential preliminary step for robotic task execution, which involves predicting grasps on regions of target objects that facilitate intended tasks. Existing literature reveals there is a limited availability of TOG datasets for training and benchmarking despite large demand, which are often synthetic or have artifacts in mask annotations that hinder model performance. Moreover, TOG solutions often require affordance masks, grasps, and object masks for training, however, existing datasets typically provide only a subset of these annotations. To address these limitations, we introduce the Top-down Task-oriented Grasping (TD-TOG) dataset, designed to train and evaluate TOG solutions. TD-TOG comprises 1,449 real-world RGB-D scenes including 30 object categories and 120 subcategories, with hand-annotated object masks, affordances, and planar rectangular grasps. It also features a test set for a novel challenge that assesses a TOG solution’s ability to distinguish between object subcategories. To contribute to the demand for TOG solutions that can adapt and manipulate previously unseen objects without re-training, we propose a novel TOG framework, Binary-TOG. Binary-TOG uses zero-shot for object recognition, and one-shot learning for affordance recognition. Zero-shot learning enables Binary-TOG to identify objects in multi-object scenes through textual prompts, eliminating the need for visual references. In multi-object settings, Binary-TOG achieves an average task-oriented grasp accuracy of 68.9\%. Lastly, this paper contributes a comparative analysis between one-shot and zero-shot learning for object generalization in TOG to be used in the development of future TOG solutions. The dataset, code, and models will be released upon publication.
\end{abstract}


\section{Introduction}
Task-oriented grasping (TOG) is a robotics field that involves predicting and executing grasps on regions of objects that facilitates the completion of a task. For instance, grasping the head of a \texttt{hammer} during a \emph{handover} task allows the user to safely retrieve the object from the handle. TOG is a critical initial step toward enabling robots to perform complex tasks, enabling advancements across multiple robotic domains such as industrial automation, domestic robotics, assistive robotics and more. A primary challenge for developing autonomous and scalable TOG solutions in such domains is ensuring that they can efficiently adapt and manipulate previously unseen objects, especially given the limited availability of TOG datasets. 

\begin{figure}
    \centering
    \includegraphics[width=\linewidth]{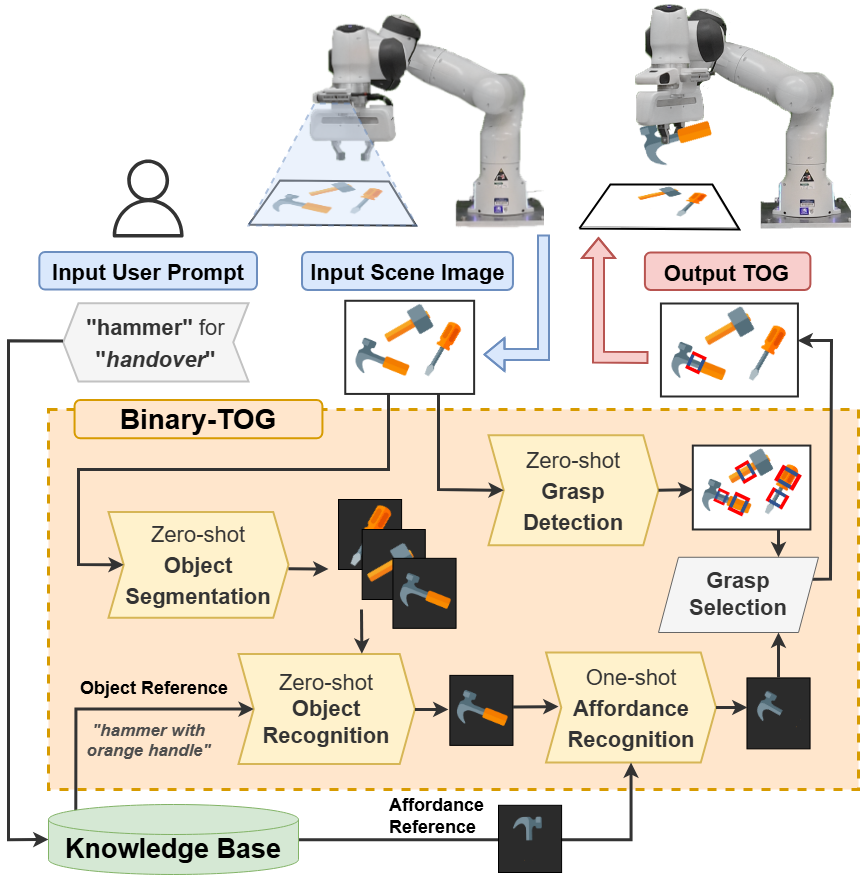}
    \caption{Binary-TOG, a framework for task-oriented grasping that predicts a task relevant grasp rectangle on an input RGB scene image that satisfies the input user prompt. The zero-shot object recognition identifies which segmented scene object is the target object based of a textual description of the object.}
    \label{fig:binarytog_simplified}
\end{figure}

To address this key challenge, we presented OS-TOG in our earlier work (\cite{holomjova_2023}), a novel TOG framework that uses one-shot learning models to generalize to both new object categories and tasks without having to be retrained. One-shot learning is a machine learning approach in which a model learns to accurately recognize or generalize when given a single labeled example. Although OS-TOG showed strong generalization, its one-shot object recognition performance was limited by the quality of its initial training datasets. In the meantime, the rise of Large Language Models (LLMs) based on foundation models (\cite{bommasani_2011}), which are large-scale architectures pre-trained on diverse datasets, resulted in the emergence of zero-shot learning solutions into TOG frameworks (\cite{tang_2023, li_2024, tang_2024}). Unlike one-shot learning, which requires a single labeled example, zero-shot learning allows models to generalize to unseen objects or tasks using only prior knowledge or semantic descriptions, such as a single textual prompt. Due to this emerging trend, this paper introduces a novel TOG framework, Binary-TOG, which modifies the one-shot object recognition module from OS-TOG to use zero-shot learning, enabling it to identify objects in multi-object scenes from textual prompts (Figure \ref{fig:binarytog_simplified}). Additionally, we compare the object generalization capabilities between OS-TOG and Binary-TOG to highlight the strengths and limitations of one-shot and zero-shot approaches through dataset and real-world experiments with a robotic manipulator. Beyond providing researchers a better understanding of each approaches current capabilities, this analysis identifies the suitability of each approach for distinct use cases and domains, and can guide future research efforts in task-oriented grasping. A key distinction between Binary-TOG and current zero-shot TOG approaches (\cite{tang_2023, tang_2024}) is that Binary-TOG decomposes task-oriented grasp selection into separate modules (i.e. object segmentation, recognition, and affordance detection). This modular design improves interpretability, facilitates error tracing, and enables clearer comparisons with one-shot learning approaches by making failure points in zero-shot methods more explainable. 

Despite the growing development of TOG solutions, the field lacks datasets for training and benchmarking, as solutions are mostly evaluated solely through real-world trials, which are prone to human error or inconsistencies between trials such as lighting or object placement. Current datasets are mostly synthetic, or have incomplete sets of annotations that are often needed for TOG (object masks, affordance masks, or grasps). Lastly, TOG datasets containing masks often contain artifacts that affects model performance and evaluation. To overcome these limitations, we present the Top-down Task-oriented Grasping (TD-TOG) dataset, a new manually annotated dataset for training and evaluating TOG solutions. TD-TOG includes highly detailed segmentation masks, visual and textual references of each object in the dataset along with semantic information (e.g. dimensions, size, weight), making it well-suited for training one-shot learning solutions. We also describe algorithms that were used to pre-label portions of TD-TOG to significantly reduce the manual effort involved in labeling. To the best of our knowledge, previous TOG literature has also largely overlooked evaluating the capabilities of TOG solutions in distinguishing between object subcategories (e.g. blue round mug, orange square mug) rather than just categories (e.g. mug, hammer). This ability is crucial in manufacturing, where small differences (e.g. bolts, screws) can significantly impact task success, or in assistive settings where recognizing personal items (e.g. a favorite cereal brand) is essential. To address this gap, we introduce a new separate challenge test set in TD-TOG that evaluates the performance of TOG solutions on differentiating between object subcategories.

The main research objective of this study is to measure the effectiveness of integrating zero-shot learning techniques to TOG solutions for generalizing to new object categories in comparison to one-shot learning techniques. The following \textbf{contributions} are made in this paper:
\begin{enumerate}
  \item Introduce a hand-annotated multi-object dataset, TD-TOG, for training and benchmarking TOG solutions. Experiments demonstrate that standard TOG solutions trained on the dataset can achieve an overall task-oriented grasp accuracy of 90.9\% on known objects and 65.9\% on new object subcategories.
  \item Present a novel TOG framework, called Binary-TOG, that leverages zero-shot and one-shot models for generalizing to new objects and tasks, respectively. In multi-object RGB scenes, Binary-TOG achieves an average task-oriented grasp accuracy of 68.9\%, without the aforementioned two models being retrained.
  \item Propose a new evaluation challenge for TOG solutions that assesses their ability to distinguish between object subcategories.
  \item Conduct real-world experiments with a robotic arm to demonstrate Binary-TOG's ability to execute TOG predictions on target objects and tasks from multi-object RGB-D scenes of household objects and tools.
\end{enumerate}

The remainder of the paper is structured as follows. The next section outlines background research and highlights research gaps in similar literature (see \nameref{sec:background_and_literature}). Next, we outline the TD-TOG dataset and its labeling methodology (see \nameref{sec:dataset}), followed by an introduction to the problem statement and the implementation of Binary-TOG, OS-TOG, and a baseline framework (see \nameref{sec:TOG_frameworks}). The remaining sections describe the experiments conducted and results obtained (see \nameref{sec:experiments}), followed by a discussion on findings and future work (see \nameref{sec:discussion}), and the conclusion (see \nameref{sec:conclusion}).

\begin{table*}
\centering
\caption{A comparison of datasets and simulated scenes used for training task-oriented grasping solutions}
\begin{threeparttable}
\resizebox{\textwidth}{!}{%
\begin{tabular}{lcccccccccccc}
\textbf{Dataset} &
  \textbf{\begin{tabular}[c]{@{}c@{}}Input\\ Modality\tnote{1}\end{tabular}} &
  \textbf{\begin{tabular}[c]{@{}c@{}}Top\\ Down\tnote{2}\end{tabular}} &
  \textbf{\begin{tabular}[c]{@{}c@{}}No. of\\ Tasks\tnote{3}\end{tabular}} &
  \textbf{\begin{tabular}[c]{@{}c@{}}No. of Obj. \\ Categories\tnote{4}\end{tabular}} &
  \textbf{\begin{tabular}[c]{@{}c@{}}No. of \\ Scenes\tnote{5}\end{tabular}} &
  \textbf{\begin{tabular}[c]{@{}c@{}}Multi\\ Object\tnote{6}\end{tabular}} &
  \textbf{\begin{tabular}[c]{@{}c@{}}Real\\ World\tnote{7}\end{tabular}} &
  \textbf{\begin{tabular}[c]{@{}c@{}}Manual\\ Annot.\tnote{8}\end{tabular}} &
  \textbf{\begin{tabular}[c]{@{}c@{}}Grasp \\ Config\tnote{9}\end{tabular}} &
  \textbf{\begin{tabular}[c]{@{}c@{}}Obj.\\ Segm.\tnote{10}\end{tabular}} &
  \textbf{\begin{tabular}[c]{@{}c@{}}Aff.\\ Segm.\tnote{11}\end{tabular}} &
  \textbf{\begin{tabular}[c]{@{}c@{}}Publicly \\ Available\tnote{12}\end{tabular}} \\ \hline
\cite{detry_2017} & D      & \cmark & 4     & 10     & 5,000          & \cmark & \xmark & \cmark & -     & \cmark & \cmark & \xmark \\
\cite{yang_2019}   & D      & \cmark & 11    & 10     & 8,000          & \cmark & \xmark & \cmark & 2D    & \xmark & \xmark & \cmark \\
\cite{fang_2020}   & D      & \cmark & 2     & 18,000 & 100,000        & \xmark & \xmark & \xmark & 2D    & \xmark & \xmark & \xmark \\
\cite{lin_2020}    & RGB-D   & \xmark & 6     & 6      & 100,000        & \cmark & \xmark & \xmark & 6-DoF & \cmark & \cmark & \cmark \\
\cite{murali_2020} & RGB-D P & \xmark & 56    & 75     & 191            & \xmark & \cmark & \cmark & 6-DoF & \xmark & \xmark & \cmark \\
\cite{liu_2020}    & RGB-D   & \xmark & 7     & 5      & 44             & \xmark & \cmark & \cmark & 6-DoF & \xmark & \xmark & \cmark \\
\cite{wen_2022}    & P      & \xmark & 2     & 3      & N/A            & \cmark & \xmark & \xmark & 6-DoF & \cmark & \cmark & \cmark \\
\cite{tang_2023}   & RGB    & \cmark & 38    & 28     & N/A            & \cmark & \xmark & \xmark & 2D    & \xmark & \xmark & \xmark \\ \hline
TD-TOG \textbf{(ours)}       & RGB-D   & \cmark & 21    & 30     & 1,449          & \cmark & \cmark & \cmark & 2D    & \cmark & \cmark & \cmark \\ \hline
\end{tabular}%
}
\begin{tablenotes}
\footnotesize
\item[1, 2] Modality of scene (D = depth, RGB-D = RGB + depth, P = point cloud) and whether scenes are top-down views.
\item[3, 4] Number of distinct tasks or object categories included.
\item[5, 6] Number of scenes (N/A if unspecified) and whether scenes are multi-object.
\item[7, 8] Whether the data was collected in the real world and annotations were manually annotated.
\item[9] The type of grasp annotations provided in dataset (2D (planar), 6-DoF, or not included in dataset (-)).
\item[10, 11, 12] Whether the dataset includes object masks or affordance masks, and is publicly available online.
\end{tablenotes}
\end{threeparttable}
\label{tab:tog_literature}
\end{table*}

\section{Background and Related Work\label{sec:background_and_literature}}
The following section includes background research on grasp detection, one-shot and zero-shot learning in the context of TOG. It then provides a review of existing literature in TOG to highlight the research gaps.

\subsection{Deep Learning in Grasp Detection}
\label{subsec:lit_grasp_detection}
Grasp detection is a field in robotics that involves predicting grasp configurations for a robotic end-effector, such as a parallel-jaw gripper or a multi-fingered hand, to achieve a stable grasp on an object based on sensor data (e.g. RGB, depth, or RGB-D images). Deep learning is one of the most widely adopted approaches for grasp detection in unstructured environments (\cite{mousavian_2019, zhao_2025}), however, these approaches often require large annotated datasets to achieve high accuracy and generalize to new objects. As a result, several grasp detection datasets have been introduced over the years for training and evaluating deep learning-based approaches (\cite{fang_2023}), some of which include scenes annotated with oriented grasp rectangles (\cite{lenz_2015, dinh_2024}), which is a popular grasp configuration for top-down grasping with parallel-jaw gripper end-effectors. A commonly used parameterization for an oriented grasp rectangle $g$ is a five-dimensional vector $g = (x, y, w, h,  \theta)$ (\cite{jiang_2011}), where $(x,y)$ denotes the center of the grasp, $w$ and $h$ represent the width and height of the grasp rectangle, and $\theta \in [-\frac{\pi}{2}, \frac{\pi}{2}]$ is the orientation angle. For parallel-jaw grippers, the height remains fixed and the width is used to adjust the opening of the grippers. It should be noted that this work is constrained to top-down grasping to reduce the research problem's complexity.

\subsection{One-shot and Zero-shot Learning in Robotics}
One-shot learning is an area of machine learning focused on enabling deep learning models to recognize objects from a single labeled example, which is highly applicable in robotics settings where collecting large amounts of training data is impractical. In a robotics challenge involving categorizing objects from cluttered bins, \cite{zeng_2022} built N-net, a one-shot learning model that had the highest accuracy in recognizing novel objects. N-net is composed of two ResNet backbones (\cite{he_2016}) with non-shared weights where one stream encodes product images using frozen ImageNet features (\cite{deng_2009}), while the other processes scene images. Similar to the triplet network architecture (\cite{schroff_2015}), N-net is trained using a set of matching and non-matching pairs, where triplet loss is used to bring matching features closer by minimizing their $\ell2$ distances while pushing non-matching ones apart. \cite{hadjivelichkov_2022} presented AffCorrs, a one-shot affordance recognition model that leverages a pre-trained DINO-ViT (\cite{caron_2021}) to extract feature descriptors from reference and scene images, where a reference image is an annotated image of a single object. Clustering and cosine similarity are used to identify regions within the scene that have similar features to a user-specified affordance (a.k.a. target affordance) in the reference image.

Zero-shot learning is a more recent research area, where deep learning models are tasked with generalizing to unseen objects or tasks using only prior knowledge or semantic descriptions, such as textual prompts. The two prominent models being used for zero-shot learning are the Contrastive Language-Image Pre-Training (CLIP) model (\cite{radford_2021}) and the Segment Anything Model (SAM) (\cite{kirillov_2023}). CLIP is a vision-language model that was trained on a large-scale dataset of image-caption pairs to learn joint representations of images and text. CLIP can be used for zero-shot image classification by comparing an image's embedding to the embeddings of labels or textual prompts, and selecting the label whose embedding is most similar to that of the image. SAM is a vision foundation model trained to achieve zero-shot generalization capabilities by being able to segment unfamiliar objects and images, without the need for additional training. Since their release, several robotics solutions have embedded them into their frameworks to generalize to novel objects. For instance, \cite{lian_2024} used CLIP as a subsequent component to AffCorrs to localize an object from a set of predicted affordances in a scene. \cite{long_2024} added an adaption module to the image encoder of SAM to improve it for underwater segmentation tasks. \cite{muttaqien_2025} combined both CLIP and SAM for identifying target products from convenience store shelves.

\subsection{Task-oriented Grasping}
Task-oriented grasping (TOG) is a challenging robotics research area that involves finding a suitable grasp on an object to fulfill a particular task (a.k.a. target task). Over the years, several deep learning and reinforcement learning solutions have been proposed to autonomously predict task-oriented grasps (\cite{fang_2020}, \cite{tang_2024}). The datasets or simulations introduced to train these solutions have been summarized in Table \ref{tab:tog_literature} in order to compare the unique properties between them.

\subsubsection{Task-oriented Grasping Datasets}
By comparing the literature provided in Table \ref{tab:tog_literature}, two main research gaps in existing TOG datasets have been identified. First, there is a notable lack of real-world multi-object TOG datasets that have been manually annotated with object and affordance segmentation masks. These datasets are in high demand, as real-world multi-object scenes closely reflect the environments in which the robots will operate. In addition to this, object masks have shown to effectively isolate objects for grasping in cluttered settings (\cite{danielczuk_2019, zhang_2024}), whilst affordance masks have shown to support novel task generalization (\cite{detry_2017,lin_2020}). Our prior findings also indicate that synthetically generated affordance and object mask annotations impacted the quality of the one-shot learning models used in OS-TOG due to them having noisy or partially-segmented masks (\cite{holomjova_2023}). 

Second, current multi-object solutions do not evaluate the model's ability to distinguish between object subcategories, which is an important component for autonomous TOG solutions used in industrial applications or domestic assistive tasks. For instance, the TOG system should be able to differentiate between different types of bolts or screws during autonomous assembly (\cite{liu_2018}). To address both of these limitations and add to the growing demand for TOG datasets, we present TD-TOG, a new dataset for training and evaluating TOG solutions with detailed affordance, object and grasp annotations.

\subsubsection{Zero-shot Learning in Task-oriented Grasping}
Given the growing interest in zero-shot approaches and the demand for TOG solutions that can generalize to novel tasks and objects, several zero-shot learning solutions have been proposed for TOG in recent years. \cite{tang_2023} proposed GraspCLIP, an end-to-end task-oriented grasping architecture that embeds CLIP for encoding visual and language inputs to produce a planar task-oriented grasp. \cite{li_2024} proposed ShapeGrasp, which uses SAM for object segmentation in single-object scenes, followed by CoACD (\cite{wei_2022}) for shape decomposition, enabling an LLM to identify task-suitable regions. \cite{palo_2024} used DINO-ViT features to match a new object with previously learned observations and transfer the associated trajectory for task execution. \cite{tang_2023} extended their previous work with FoundationGrasp (\cite{tang_2024}), a method that predicts 6-DoF grasps based on an input point cloud, RGB image, and textual prompt. It leverages their extended version of the TaskGrasp dataset (\cite{murali_2020}), enriched with semantic, geometric, and web-retrieved visual data. During inference, features are extracted from multiple models (\cite{radford_2021}, \cite{mousavian_2019}, \cite{qi_2017}) and used by a TOG evaluator to select the most suitable grasp candidate.

Despite recent advances in zero-shot TOG methods, their object generalization capabilities remain underexplored compared to one-shot approaches, even though both show similar generalization to novel tasks and objects. To this end, we build upon OS-TOG and propose Binary-TOG, which adopts the same architecture as OS-TOG but uses zero-shot learning to generalize to novel object categories. A key distinction between Binary-TOG and \cite{tang_2023, tang_2024}'s approach is that Binary-TOG decomposes task-oriented grasp selection into separate modules (i.e. object segmentation, recognition, and affordance detection). This modular design improves interpretability, facilitates error tracing, and enables clearer comparisons with one-shot learning approaches by making failure points in zero-shot methods more explainable. 
\section{Top-Down Task-Oriented Grasping Dataset} 
\label{sec:dataset}
This section introduces the Top-Down Task-Oriented Grasping (TD-TOG) dataset, detailing its structure followed by the data collection and annotation process. 

\begin{figure}[!htbp]
  \centering
  \includegraphics[width=\linewidth]{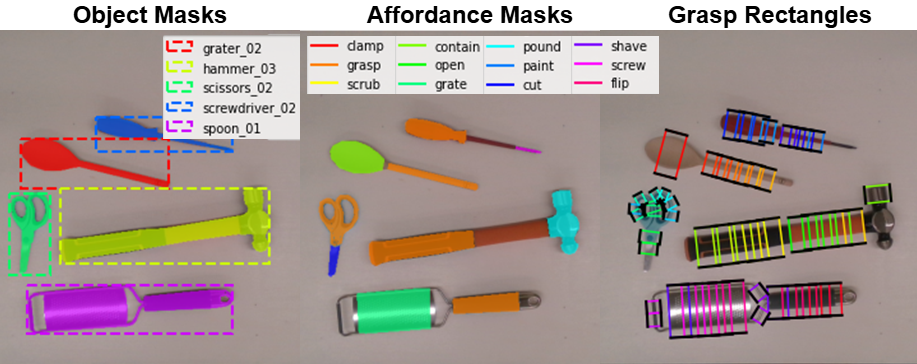}
  \caption{An example scene from the TD-TOG training set annotated with object masks (left), affordances (middle), and grasp rectangles (right).}
  \label{fig:annotations}
\end{figure}

\begin{figure*}
      \centering
  \includegraphics[width=1\linewidth]{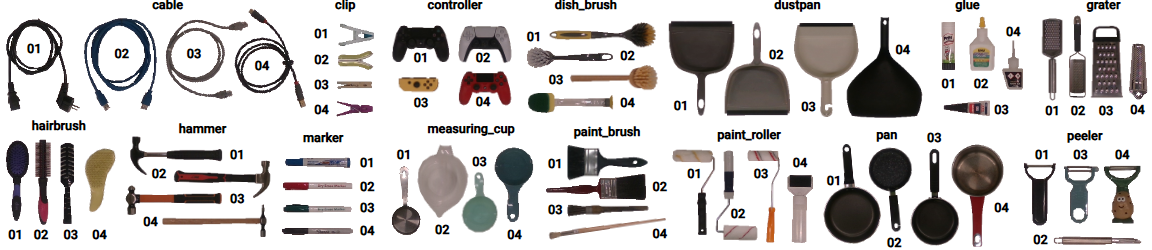}
  \caption{A preview of 15 object categories (e.g. \texttt{cable}, \texttt{clip}) and 60 object subcategories  (e.g. \texttt{cable\_01}, \texttt{cable\_02}) from the TD-TOG dataset.}
  \label{fig:dataset_objects}
\end{figure*}

TD-TOG contains 1,449 labeled RGB-D scenes of household objects intended for training and evaluating TOG systems. Each object is hand-annotated with object masks, affordance masks, grasp rectangles (Figure \ref{fig:annotations}), and textual descriptions (e.g. \textit{round thin paintbrush with wooden handle and light bristles}). TD-TOG differs from existing TOG datasets due to the following unique features: it is built from real-world data, has multi-object scenes, and includes diverse object coverage. Additionally, it provides highly detailed annotations such as precise object and affordance masks that even account for holes within objects (e.g. the handles of scissors), which to the best of our knowledge is not featured in existing TOG or affordance datasets. 

\subsection{Dataset Statistics}
\begin{figure}
  \centering
  \includegraphics[width=\linewidth]{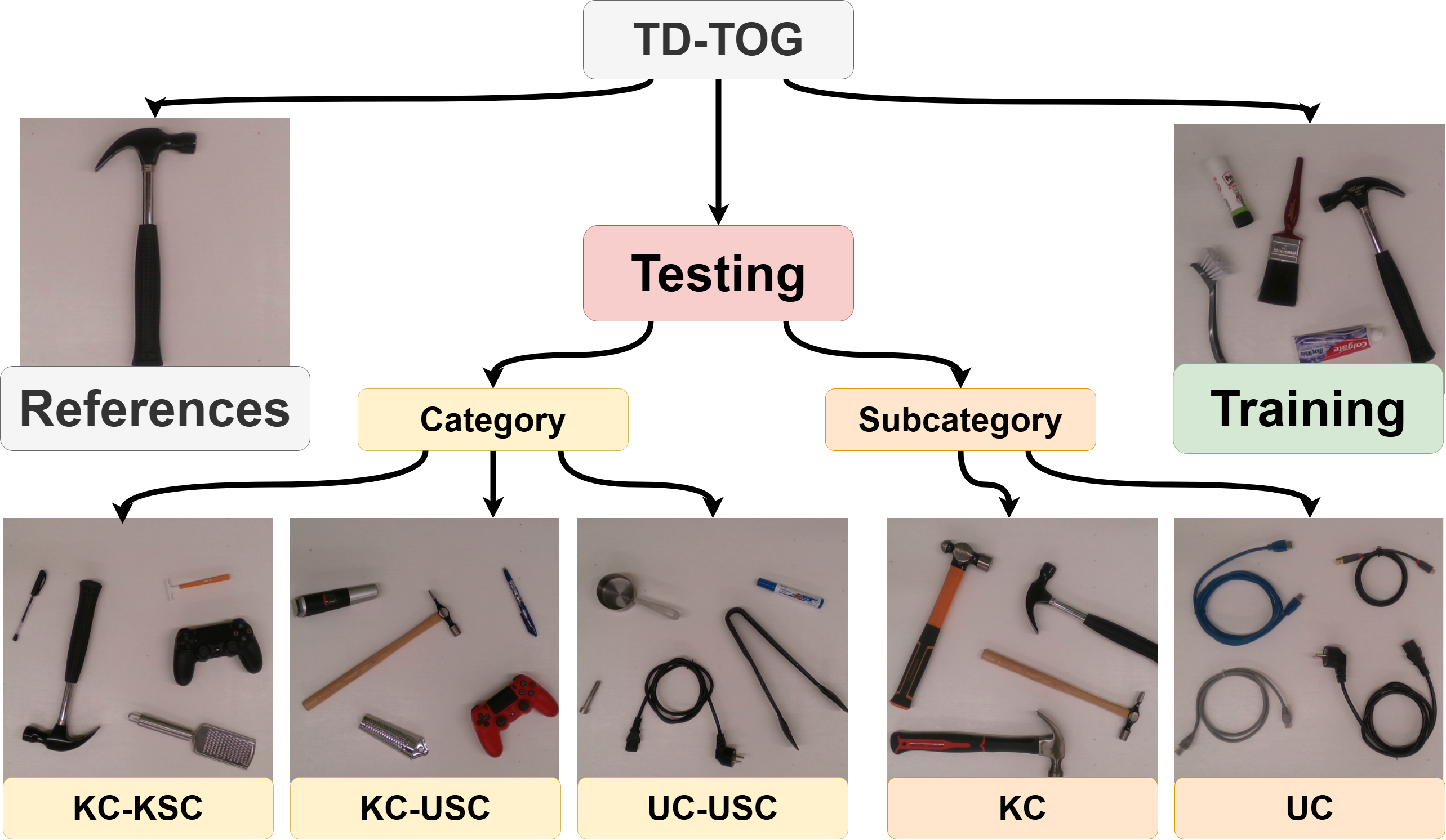}
  \caption{The structure of the TD-TOG dataset, which consists of object reference images (white), training images (green), and testing images (red). The testing set is further divided into the \textit{category} and the \textit{subcategory} splits.}
  \label{fig:dataset_hierarchy}
\end{figure}

The TD-TOG dataset features 30 unique object categories (i.e. \texttt{hammer, paintbrush}), 120 object subcategories (i.e. \texttt{hammer\_01, hammer\_02}), and 16 affordances. For each object category, there are 4 object subcategories which are variations of the same category. Furthermore, affordances can be mapped to tasks to define task-relevant regions in objects as exemplified in Table \ref{tab:task_definitions}. In total, the dataset has 4,101 annotated object instances including 31,248 grasp rectangles and 8,337 affordance masks. TD-TOG is divided into 3 main sets: references, training, and testing (Figure \ref{fig:dataset_hierarchy}). The testing set further branches into the \textit{category} and \textit{subcategory} splits.

\subsubsection{Reference and Training Sets}
The reference set has 120 single-object scenes, providing an annotated example of each subcategory in the dataset. The training set contains 750 scenes and features 15 object categories, which are referred to as known categories (KC) and are listed in Table \ref{tab:tdtog_overall_kc_results}. From this set of KC objects, the training set uses subcategories having a suffix of \texttt{\_01}, \texttt{\_02,} or \texttt{\_03}, resulting in 45 subcategories being used in the training set. To train TOG systems at varying levels of complexity, each training scene has between 1 to 5 object instances where each object has a distinct category.

\subsubsection{Category Testing Split}
The \textit{category} split has 225 scenes, and is intended for evaluating the general performance of TOG systems as carried out in existing TOG literature. Similar to the training set, the scenes comprise of 1 to 5 object instances, where each object has a distinct category. The \textit{category} split has 3 subsplits: KC-KSC, KC-USC, and UC-USC to challenge the object generalization capabilities of TOG systems at varying levels of novelty. The KC-KSC subsplit features 15 subcategories that are present in the training set, referred to as known subcategories (KSC). These 15 subcategories are KC objects that end with \texttt{\_01}. On the other hand, KC-USC and UC-USC contain subcategories that were not present in the testing set, referred to as unknown subcategories (USC). The KC-USC subsplit uses subcategories of KC objects that end with \texttt{\_04}. The UC-USC subsplit features 15 subcategories of unknown category (UC) objects (see Table \ref{tab:tdtog_overall_uc_results}), that end with \texttt{\_01}.

\subsubsection{Subcategory Testing Split}
The \textit{subcategory} split has 354 scenes and is intended for challenging TOG solutions on their ability to differentiate between object subcategories, which to our best knowledge has not been carried out in previous TOG literature. These scenes are composed of 2-4 object instances, all having the same category but different subcategories. The KC split uses all 60 subcategories of the 15 object categories that were featured in the training set (Table \ref{tab:tdtog_overall_kc_results}). The UC split features all 60 subcategories of the 15 object categories that were not featured in the training set (Table \ref{tab:tdtog_overall_uc_results}).

\subsection{Data Collection}
\begin{figure}[!htbp]
  \centering
  \includegraphics[width=0.8\linewidth]{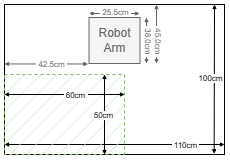}
  \caption{The experimental setup for data collection having a $110{\times}100\textrm{cm}$ white table with a fixed $25.5{\times}38.0\textrm{cm}$ metallic base on which the robotic arm is placed. The green $60{\times}50\textrm{cm}$ region of the table is restricted for object placement, where the robotic arm will grasp the object.}
  \label{fig:experimental_setup}
\end{figure}

The dataset objects were selected by combining common household items found in task-oriented grasping datasets, as well as less frequently observed items that exist in various forms (e.g. paintbrushes, graters). Figure \ref{fig:dataset_objects} shows a preview of the objects found in the dataset. Additional selection criteria for the objects included possessing at least one affordance, fitting within the experimental setup's size constraints, and encompassing a diverse range of shapes, colors, and materials. These selected objects were acquired from workshops, households, and homeware stores. RGB-D scenes of $480{\times}640\textrm{px}$ were captured using a 7-DoF robotic arm by Franka Emika equipped with a D415 Intel RealSense camera. Figure \ref{fig:experimental_setup} shows the experimental setup used for capturing object scenes.

\begin{table}
\caption{The tasks considered in this work, along with their associated affordance-based constraints and the applicable objects}
\resizebox{0.97\linewidth}{!}{%
\begin{tabular}{lll}
\textbf{Task}       & \textbf{Rule Definition} & \textbf{Affected Object Categories}   \\ \hline
\textit{transport}  & \texttt{-}                  & all objects                 \\
\textit{handover}   & \textbf{NOT} \texttt{grasp}            & all objects except cable    \\
\textit{brushing}   & \texttt{grasp}                & hairbrush, toothbrush       \\
\textit{clamping}   & \texttt{grasp}                & clip, tongs                 \\
\textit{connecting} & \texttt{connect}              & cable                       \\
\textit{cutting}    & \texttt{grasp}                & pizza cutter, scissors     \\
\textit{flipping}   & \texttt{grasp}                & spatula                     \\
\textit{frying}     & \texttt{grasp}                & pan                         \\
\textit{gluing}     & \texttt{grasp}                & glue                        \\
\textit{grating}    & \texttt{grasp}                & grater                      \\
\textit{hitting}    & \texttt{grasp}                & hammer, tenderizer          \\
\textit{measuring}  & \texttt{grasp}                & thermometer                 \\
\textit{opening}    & \texttt{open}                 & toothpaste, vitamin         \\
\textit{painting}   & \textbf{NOT} \texttt{paint}            & paint brush, paint roller \\
\textit{peeling}    & \texttt{grasp}                & peeler                      \\
\textit{scooping}   & \texttt{grasp}                & measuring cup, spoon       \\
\textit{screwing}   & \textbf{NOT} \texttt{screw}            & screw, screwdriver          \\
\textit{scrubbing}  & \texttt{grasp}                & dish brush                 \\
\textit{shaving}    & \texttt{grasp}                & razor                       \\
\textit{sweeping}   & \textbf{NOT} \texttt{contain}          & dustpan                     \\
\textit{writing}    & \texttt{grasp}                & marker, pen                 \\ \hline
\end{tabular}%
}
\label{tab:task_definitions}
\end{table}

\subsection{Data Annotation} 
Two open-source tools were used for manually annotating the dataset: Labelme (\cite{kentaro_2016}) for annotating object and affordance masks, and LabelImg2 (\cite{chinakook_2018}) for oriented grasping rectangles. To significantly reduce human annotation effort, we develop two algorithms to pre-label parts of the dataset, which are then refined manually. These algorithms are further explained below and will be released as part of the dataset through the following link\footnote{\url{https://github.com/valerija-h/td_tog}}. The remainder of this section details the annotation process.

The annotation process begins with manually annotating detailed object masks and grasp rectangles for each object in the reference set. Next, we label rough affordance mask annotations consisting of simple shapes, such as squares and rectangles to reduce annotation effort. These rough masks are then refined using our first algorithm, the affordance-refining algorithm, which extracts the portion of each affordance that overlaps with the detailed object mask. The resulting refined affordance masks are subsequently reviewed by a human annotator to maintain high data quality.

The second stage of the annotation process involves annotating the testing set and training set with detailed object masks. We then apply our second algorithm, the auto-labeling algorithm, which leverages the annotations of the reference dataset to label grasp rectangles and rough affordance masks on each object in each scene. This pre-labeling process significantly reduces the human annotation effort required and minimizes the potential for human error. These labels are then validated in a manual effort to ensure data quality. Similar to before, the affordance-refining algorithm is used to obtain detailed affordance masks. Once complete, the final annotations are represented in a single JSON file for each dataset split, extending the format of the COCO object detection datasets (\cite{tsung-yi_2014}).

\section{Task-Oriented Grasping Frameworks}
\label{sec:TOG_frameworks}
 This section formalizes task-oriented grasping and provides details on the TOG frameworks we developed with particular emphasis on the newly proposed framework, Binary-TOG.
 
\subsection{Problem Statement}
\label{subsec:problem_statement}
Task-oriented grasping (TOG) involves identifying a grasp configuration on a target object that enables the successful execution of a specified task. This subsection is dedicated to formally defining TOG in the context of our work.

Assume a set of tasks $T$, a set of objects $O$, and a set of affordances $A$. Each task $t_i \in T$ can be mapped to an affordance $a_j \in A$ by a rule $R$. The rule defines the affordance region that accomplishes $t_i$ (i.e. $R(t_i) = a_j$) or the region to avoid (i.e. $R(t_i) = \neg\,a_j$) as exemplified in Table \ref{tab:task_definitions}. Given as input an RGB image $\hat{I} \in \mathbb{R}^{640{\times}480{\times}3}$ containing a subset of objects $\bar{O} \subseteq {O}$, a target object $\hat{o} \in \bar{O}$ and a target task $\hat{t} \in T$, task-oriented grasping involves identifying a grasp configuration $\hat{g}^{*}$ on the object $\hat{o}$ for the task $\hat{t}$ that satisfies the conditions defined by $R(\hat{t})$. The grasp configuration $\hat{g}^{*}$ is equivalent to the configuration $g$ previously defined (see \nameref{subsec:lit_grasp_detection}).

\subsubsection{Knowledge Base}
In the case of one-shot and binary-shot TOG, we define a knowledge base $K$ for every object $o \in O$, where each entry $k_o$ for the object $o$ is defined as:
\[k_{o} = (I_{o}, M_{o}, F_{o}, G_{o}, D_{o})\]
\begin{conditions*}
I_o  &  RGB image of the object $I_o \in \mathbb{R}^{640{\times}480{\times}3}$ \\
M_o  &  binary object mask $M_o \in \{0,1\}^{640{\times}480}$, where 0 is the background and 1 is the object $o$. \\
F_o  &  binary affordance masks $F_o = \{F_o^{a} \mid a \in A_i\}$, where $A_i \subseteq A$ is the subset of affordances applicable to object $o$, and $F_o^{a} \in \{0,1\}^{640{\times}480}$ is the mask for affordance $a$. \\
G_o  &  grasp rectangles $G_o = \{g \mid g = (x, y, w, h, \theta)\}$ for object $o$. \\
D_o  &  textual description of the object $o$ as a sentence
\end{conditions*}

\begin{figure*}
    \centering
    \includegraphics[width=1\linewidth]{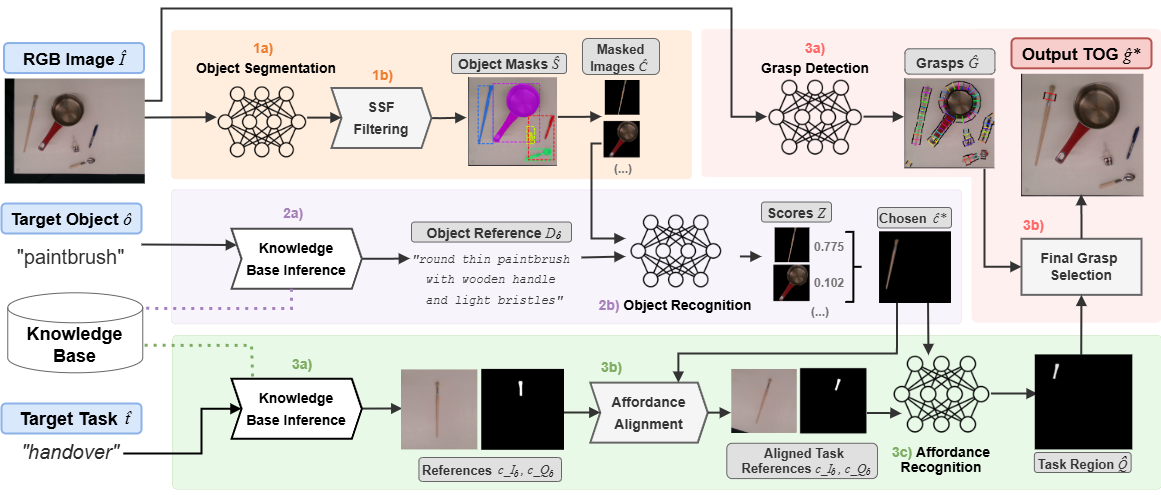}
    \vspace*{-7mm}
    \caption{The architecture of Binary-TOG, which is composed of four modules: zero-shot object segmentation (orange), zero-shot object recognition (purple), one-shot affordance recognition (green), and zero-shot grasp detection (red).}
    \label{fig:binarytog_architecture}
\end{figure*}

\subsection{Binary-TOG Framework}
Binary-shot Task-Oriented Grasping (Binary-TOG) is a novel TOG framework that can generalize to new objects and tasks without having to re-train any of its underlying submodels. This generalization is achieved via embedded one-shot and zero-shot models that leverage a single visual or textual reference of the target object from a knowledge base $K$ during inference.
By adding entries to $K$, Binary-TOG can readily recognize and adapt to new tasks and objects. The term \textit{binary-shot} reflects the framework's two main modules responsible for its generalization capabilities: the \textit{zero-shot} object recognition module that recognizes new objects, and the \textit{one-shot} affordance recognition module which generalizes to novel tasks that can be performed with those objects through logical mappings to affordances.

As shown in Figure \ref{fig:binarytog_architecture}, Binary-TOG consists of four embedded neural networks, supported by arrow-shaped processing blocks and knowledge base (KB) retrieval blocks. The processing blocks are preprocessing or postprocessing algorithms designed to improve model performance, execute decision-making, and enable the flow of data between models. The KB retrieval components obtain references from $K$ during inference. Refer to \nameref{subsubsec:model_selection}, for further details on which models were selected for each module in Binary-TOG for the final evaluation.

\begin{algorithm}
\caption{Size Subset Filtering (SSF)}
    \begin{algorithmic}[1]
        \Require $\hat{I}, S, min\_area, max\_area, \tau$
        \Ensure $\hat{S}$
        \State $\hat{S} \gets S$
        \ForAll{$seg \in S$}
            \State $seg\_area \gets$ sum($seg$)
            \If{not $min\_area < seg\_area < max\_area$}
                \State $\hat{S}$.remove($seg$)
            \EndIf
        \EndFor
        
        \State $keep\_idxs \gets [ ]$
        \For{$i \gets 0$ to length($\hat{S}$)}
            \State $to\_remove \gets 0$
            \For{$j \gets i+1$ to length($\hat{S}$)}
                \State $m \gets \hat{S}[i] \land \hat{S}[j]$
                \State $p \gets $ sum($m$) \verb|/| sum($\hat{S}[i]$)
                \If{$p > \tau$}
                    \State $to\_remove \gets 1$
                    \State \textbf{break}
                \EndIf
            \EndFor
            \If{not $to\_remove$}
                \State $keep\_idxs$.add($i$)
            \EndIf
        \EndFor
        \State $\hat{S} \gets \hat{S}[keep\_idxs]$
    \end{algorithmic}
\label{algorithm:size_subset_filtering}
\end{algorithm}

\subsubsection{Zero-shot Object Segmentation Module}
\label{subsec:zeroshot_object_segmentation}
The initial module of the framework uses a zero-shot segmentation model to segment regions within the input image scene $\hat{I}$, generating a set of binary segmentation masks $S \subseteq \{0,1\}^{640{\times}480}$. However, this model can produce an excessive number of segmentations, including background regions and multiple overlapping object fragments. Thus, to reduce distractions and enhance the performance of the subsequent image recognition module, we introduce a filtering algorithm called Size Subset Filtering (SSF), which filters $S$ to a subset $\hat{S} \subseteq S$ to only retain segmentations that represent objects in the scene (Algorithm \ref{algorithm:size_subset_filtering}).

As defined in Algorithm \ref{algorithm:size_subset_filtering}, SSF operates in two distinct stages. The first stage discards segmentation masks that deviate from predefined size thresholds, $min\_area$ and $max\_area$, ensuring the elimination of small noise artifacts and large background regions. These thresholds were empirically set to 400\textrm{px}$^2$ and 50,000\textrm{px}$^2$, respectively, to maintain an optimal balance between retaining meaningful objects whilst removing irrelevant segments. The second stage applies further filtering by identifying and removing redundant subset masks. This is achieved by computing the overlap percentages between the masks and eliminating those that exceed the predefined overlap threshold $\tau$, which was found to yield optimal results at $\tau = 0.75$.

\subsubsection{Zero-shot Object Recognition Module}
The second module starts by retrieving a textual description $D_{\hat{o}}$ of the target object $\hat{o}$ from its knowledge base $K$. It then applies each binary segment mask from $\hat{S}$ on the input scene image $\hat{I}$ to obtain a set of masked images $\hat{C}$, defined as:
\[ \hat{C} = \{\hat{c}_i \in \mathbb{R}^{640{\times}480{\times}3} \,|\, \hat{c}_i = \hat{I} \cdot\hat{s}_i, \: \forall \hat{s}_i \in \hat{S}\}\]

The inputs $\hat{C}$ and $D_{\hat{o}}$ are then passed into a zero-shot object recognition model which outputs feature vectors that are normalized to facilitate cosine similarity computation. The normalized image features of $\hat{C}$ are multiplied by the transposed normalized textual feature of $D_{\hat{o}}$, and scaled by a factor of 100 to highlight score differences. These scores are passed through a softmax function to obtain the confidence scores $Z = (z_1, \dots, z_N)$, where each score $z_i \in [0,1]$ represents the confidence that a masked image $\hat{c}_i$ corresponds to $D_{\hat{o}}$, which can be expressed as:
\[z_i = \text{model($\hat{c}_i$, $D_{\hat{o}}$)}, \: \forall \hat{c}_i \in \hat{C}\] 

Given that there are $N$ images in $\hat{C}$, a postprocessing block selects the masked image with the highest confidence score, which we define as $\hat{c}^{*}$. The selected masked image $\hat{c}^{*}$ is now assumed to represent the target object $\hat{o}$ as it had the highest correspondence confidence with its description $D_{\hat{o}}$.
\[\hat{c}^{*} = \hat{c}_k \quad\textrm{where}\quad k = \argmax_{i \in \{1,...,N\}} z_i\]

\subsubsection{One-shot Affordance Recognition Module}
The knowledge base retrieval block starts by determining the affordance $\hat{a}$ associated with target task $\hat{t}$ through the rule $R(\hat{t})$. This is used to retrieve the affordance mask $F^{\hat{a}}_{\hat{o}}$, the object mask $M_{\hat{o}}$ and the object RGB image $I_{\hat{o}}$ corresponding to the target object $\hat{o}$ from the knowledge base. Depending on the polarity of the association of the rule $R(\hat{t})$ with the affordance $\hat{a}$, a binary mask $Q_{\hat{o}} \in \{0,1\}^{640{\times}480}$ can be obtained as follows:
\[
Q_{\hat{o}} = 
\begin{cases}
    M_{\hat{o}} \cap F_{\hat{o}}^{\hat{a}}, & \text{if } R(\hat{t}) = \hat{a}\\
    M_{\hat{o}} \setminus F_{\hat{o}}^{\hat{a}}, &  \text{if } R(\hat{t}) = \neg\, \hat{a}\\
    M_{\hat{o}}, & \text{otherwise}
\end{cases}
\]
Thus, the reference task-suitable region $Q_{\hat{o}}$ is a binary mask denoting the region of $I_{\hat{o}}$ where grasps need to be situated to accomplish $\hat{t}$. In the case where no rules are specified for a task (e.g. \textit{transport}), the object mask $M_{\hat{o}}$ is used and the model inference step is skipped.

Prior to model inference, $I_{\hat{o}}$, $Q_{\hat{o}}$ and $\hat{c}^{*}$ are cropped using their object's bounding box and padded to $256{\times}256\textrm{px}$, resulting in $c\_I_{\hat{o}}$, $c\_Q_{\hat{o}}$ and $c\_\hat{c}^{*}$. Additionally, we introduce an affordance alignment (AA) module that rotates $c\_I_{\hat{o}}$ and $c\_Q_{\hat{o}}$, so that they align with the orientation of the object in $c\_\hat{c}^{*}$ (Algorithm \ref{algorithm:affordance_alignment}). This module iterates through a specified number of rotations $n\_rots$, and determines the rotation that minimizes the squared L2 norm between $c\_Q_{\hat{o}}$ and the masked image $c\_I_{\hat{o}} \cdot c\_\hat{c}^{*}$. This optimal rotation is then applied to both $c\_I_{\hat{o}}$ and $c\_Q_{\hat{o}}$.

\begin{algorithm}
\caption{Affordance Alignment (AA)}
    \begin{algorithmic}[1]
        \Require $c\_\hat{c}^{*}, c\_I_{\hat{o}}, c\_Q_{\hat{o}}, n\_rots$
        \Ensure $c\_I_{\hat{o}}, c\_Q_{\hat{o}}$
        
        \State $c\_c_{\hat{o}}$ = $c\_I_{\hat{o}} \cdot c\_Q_{\hat{o}}$ 
        \State $min\_dist$ = $||c\_\hat{c}^{*} - c\_c_{\hat{o}}||^2$
        \State $min\_r$ = 0
        \For{$i \gets 0$ to $n\_rots$}
            \State $r =  0 + (i*(360//n\_rots))$
            \State $rot\_c_{\hat{o}} = $ rotate($c\_c_{\hat{o}}$, $r$)
            \State $dist = ||c\_\hat{c}^{*} - rot\_c_{\hat{o}}||^2$
            \If{$dist < min\_dist$}
                \State $min\_dist = dist$
                \State $min\_r = r$
            \EndIf
        \EndFor

        \If{$min\_r \neq 0$}
            \State $c\_I_{\hat{o}} = $ rotate($c\_I_{\hat{o}}$, $min\_r$)
            \State $c\_Q_{\hat{o}} = $ rotate($c\_Q_{\hat{o}}$, $min\_r$)
        \EndIf
    \end{algorithmic}
\label{algorithm:affordance_alignment}
\end{algorithm}

A one-shot affordance recognition model takes as input $c\_\hat{c}^{*}$, along with the reference cropped object image $c\_I_{\hat{o}}$ and its corresponding affordance mask $c\_Q_{\hat{o}}$. It then outputs a single binary mask $c\_\hat{Q}$ denoting the task-suitable region in $c\_\hat{c}^{*}$. Subsequently, $c\_\hat{Q}$ is unpadded and uncropped as $\hat{Q}$ to follow the original proportions of $\hat{c}^{*}$. Therefore, the binary mask $\hat{Q} \in \{0,1\}^{640{\times}480}$ identifies the regions in $\hat{c}^{*}$ where grasps must be positioned to accomplish task $\hat{t}$.

\subsubsection{Zero-shot Task-Oriented Grasp Module}
Given the original input image $\hat{I}$, a grasp detection model generates a set of candidate grasps $\hat{G}$, where each grasp $\hat{g} \in \hat{G}$ is equivalent to the definition of $g$ (see \nameref{subsec:lit_grasp_detection}), with an extra parameter $z \in [0,1]$ denoting the model's confidence:
\[\hat{G} = \{\hat{g}_j \,|\ \hat{g}_j = (x, y, w, h, \theta, z)\}\] 
It then filters $\hat{G}$ to retain task-relevant grasps by keeping those having grasp coordinates $(x,y)$ that fall within the task-suitable region (i.e. where $\hat{Q}[x, y] = 1$). From these filtered candidates, the grasp with the highest confidence is selected as the final task-oriented grasp $\hat{g}^{*}$.

\subsection{OS-TOG Framework (Previous Work)}
\label{subsec:ostog_framework}
OS-TOG is a TOG framework developed in our previous work (\cite{holomjova_2023}), having a different architecture to Binary-TOG, but maintaining similar generalization capabilities towards new objects and tasks by leveraging its knowledge base $K$. The major distinction is the object recognition module of OS-TOG, which uses a one-shot learning model that depends on visual references of the target object rather than textual references. Additionally, the OS-TOG in this paper adopts the improved object segmentation module with the added SSF module described in \nameref{subsec:zeroshot_object_segmentation}. The underlying model remains category-agnostic and maintains the same input and output formats as before, but the combination of a foundation model with the SSF module produces improved results in real-world scenes.

\subsubsection{One-Shot Object Recognition Module}
\label{subsec:oneshot_object_recognition}
Similar to Binary-TOG, a set of masked images $\hat{C}$ is built by applying each binary segment mask $\hat{s} \in \hat{S}$ from the previous object segmentation module to the input scene image $\hat{I}$. Next, the module retrieves a reference RGB image $I_{\hat{o}}$ and an object mask $M_{\hat{o}}$ of the target object $\hat{o}$ from its knowledge base $K$. The reference object mask $M_{\hat{o}}$ is applied to $I_{\hat{o}}$ to obtain a reference masked image $c_{\hat{o}} = I_{\hat{o}} \cdot M_{\hat{o}}$ of the target object $\hat{o}$. Cropping and padding is applied to $c_{\hat{o}}$ and each image in $\hat{c} \in \hat{C}$ to size of $256{\times}256\textrm{px}$, resulting in $c\_c_{\hat{o}}$ and $c\_\hat{C}$, respectively. This step ensures that objects are magnified to a comparable size, which facilitates the similarity comparison of small-sized objects.

The preprocessed inputs $c\_c_{\hat{o}}$ and $c\_\hat{C}$ are fed into a one-shot learning model, which generates a feature vector $f\_c_{\hat{o}}$ and a set of feature vectors $f\_\hat{C}$. A postprocessing step determines which feature $f\_\hat{c}_k \in f\_\hat{C}$ has the minimum $\ell2$ distance to $f\_c_{\hat{o}}$. The corresponding masked image of the feature, denoted as $\hat{c}^*$, is then selected and assumed to best represent the target object $\hat{o}$.
\[\hat{c}^{*} = \hat{c}_k \quad\textrm{where}\quad k = \argmax_{k \in \{1, ... ,N\}} ||f\_\hat{c}_k - f\_c_{\hat{o}}||_2\]

The masking stage is essential for background removal and eliminating distractions (e.g. another object appearing in the same frame). Since, there is a lack of pre-trained models that produce meaningful feature vectors from masked-out backgrounds, the one-shot learning model needs preliminary training. However, this would only be carried out once and is readily scaled to new objects after. Further details on training and the one-shot learning model used is provided in \nameref{subsubsec:model_selection}.

\subsection{Standard-TOG Framework}
Standard-TOG is a TOG framework we introduce in this work to use as a performance baseline for both OS-TOG and Binary-TOG. While it adopts the same overall architecture as Binary-TOG, it has different object recognition and affordance recognition modules. The underlying models in these modules are standard neural networks that do not rely on a KB and are trained to recognize a fixed set of object and affordance categories. This eliminates the dependency on the KB at the cost of reduced generalization capabilities. Thus, this framework is not able to participate in trials involving the recognition of new objects or tasks. Standard-TOG was designed to imitate the performance and capabilities of traditional TOG solutions that do not incorporate one-shot or zero-shot techniques.

\subsubsection{Standard Object Recognition Module}
\label{subsec:standard_object_recognition}
As carried out in previous frameworks, a set of masked images $\hat{c}_i \in \hat{C}$ is formed and then cropped and padded to a size of $256{\times}256\textrm{px}$, resulting in $c\_\hat{C}$. Given the temporary assumption there are $N$ images in $\hat{C}$, and $M$ categories that the standard object recognition is trained on. When passing $c\_\hat{C}$ as input, the standard object recognition produces a logit matrix $Y \in \mathbb{R}^{N \times O}$ which is passed through a softmax function to obtain a probability matrix $Z \in \mathbb{R}^{N \times O}$, where $z_{ij}$ is the probability that $\hat{c}_i$ corresponds to $j^{th}$ object category. To identify the masked image $\hat{c}^{*}$ that most represents the target object $\hat{o}$, we select the image corresponding to the highest probability in the column associated with $\hat{o}$.

\subsubsection{Standard Affordance Recognition Module}
\label{subsec:standard_affordance_recognition}
The module first determines the affordance $\hat{a}$ associated with the rule $R(\hat{t})$, given that $\hat{t}$ is the target task. In the case of no affordance specified in the rule, the model inference process is skipped and $\hat{c}^{*}$ is returned. Next, it preprocesses $\hat{c}^{*}$ by cropping it and padding it to a size of $256{\times}256\textrm{px}$, resulting in $c\_\hat{c}^{*} \in \mathbb{R}^{256 \times 256 \times 3}$. The affordance recognition model, which we chose to use an instance segmentation model as it produces multiple affordance and allows multiple affordances without having to separate them. This model will produce a set of binary affordance masks $c\_\hat{F}$ and predicted affordance labels $\hat{L}$. The cropped binary affordance masks are resized to the original size of $640{\times}480\textrm{px}$ as $\hat{F}$. We then stack all predictions above a confidence threshold that have the affordance label of $\hat{a}$ and generate the task-suitable region $\hat{Q}$ as follows:

\[
\hat{Q} = 
\begin{cases}
    \hat{c}^{*} \cap \hat{F}, & \text{if } R(\hat{t}) = \hat{a}\\
    \hat{c}^{*} \setminus \hat{F}, &  \text{if } R(\hat{t}) = \neg\, \hat{a}\\
    \hat{c}^{*}, & \text{otherwise}
\end{cases}
\]

\section{Experiments and Results}
\label{sec:experiments}
This section provides further context on how the frameworks were implemented and evaluated, followed by presenting the results obtained during experiments on both TD-TOG and in real-world experiments.

\subsection{Implementation Details}
The frameworks are implemented using Python 3.9, and any models are trained using PyTorch 2.0.1 with CUDA 11.7. Real-world experiments are carried out using the same experimental setup shown in Figure \ref{fig:experimental_setup}, where the depth image is used to calculate the distance between the object and camera to execute the predicted task-oriented grasp. The Frankx library (\cite{frankx}) was used for motion planning.

\subsubsection{Model Selection} \label{subsubsec:model_selection}
The following models were used in Binary-TOG, OS-TOG, and Standard-TOG for this study as they yielded optimal results in preliminary trials under the resources previously specified. The Segment Anything Model (SAM) (\cite{kirillov_2023}) with the ViT-L backbone is used for zero-shot object segmentation. The S$^{2}$A-Net (\cite{jiaming_2022}) is trained on TD-TOG for grasp detection as carried out in previous work (\cite{holomjova_rotated}). One-shot affordance recognition used by both Binary-TOG and OS-TOG is carried out by AffCorrs (\cite{hadjivelichkov_2022}). In the case of Binary-TOG, The ViT-B/32 variant of the Contrastive Language-Image Pretraining (CLIP) (\cite{radford_2021}) is used for zero-shot object recognition. OS-TOG uses a re-implementation of N-net (\cite{zeng_2022}) for one-shot object recognition. In the case of OS-TOG, standard object segmentation is carried out by training a ResNet-50 (\cite{he_2016}) on TD-TOG. Lastly, standard affordance recognition is carried out using a Mask R-CNN (\cite{kaiming_2017}) model trained on TD-TOG. Additional model configurations, data augmentations, and training parameters will be released through the following link\footnote{\url{https://github.com/valerija-h/binary_tog}} upon publication.

\subsection{Evaluation Metrics}
The frameworks are evaluated on their performance in TD-TOG and real-world experiments. On TD-TOG, they are evaluated on the \textit{category} split to assess their general performance on TOG and the \textit{subcategory} split to challenge the object recognition components in differentiating between object subcategories. Real-world experiments are carried out to showcase the predictions on a robotic platform.

\subsubsection{Dataset Experiments}
In experiments on the \textit{category} split, the frameworks are assessed by iterating through each scene in the split, and then each object in the scene and its associated tasks (see Table \ref{tab:task_definitions}). Therefore, a scene of 2 objects having 3 tasks each would generate 6 TOG predictions per framework. Each TOG prediction contains the predicted task-oriented grasp and also contains predictions of object segmentation, affordance recognition, and standard grasp detection. For each trial, the performance of each of these modules is recorded using gold-standard metrics from literature. 

Object segmentation performance is evaluated using standard COCO dataset metrics (\cite{lin_2020}), including the mean Average Precision (AP) and the Intersection over Union (IoU) score. Object recognition is calculated using the $F_1$ score, as well as affordance recognition since it was infeasible to compute the $F^w_b$ score (\cite{margolin_2014}) given the size of the affordance mask. Grasp accuracy is calculated by classifying a predicted grasp as a success if it has an IoU score greater than 25\% with a ground-truth grasp and a $\theta$ angle difference within 30$^{\circ}$. This is extended to Task-oriented Grasp (TG) accuracy by also checking if the center point of grasp $(x,y)$ is in the ground-truth task-relevant region. In experiments on the \textit{subcategory} split, only the performance of the object recognition modules is presented showing their accuracy and $F_1$ score.

\subsubsection{Real-world Experiments}
In real-world experiments, we evaluate the performance on novel objects from the KC-USC and UC-USC subsplits. Experiments are carried out by creating 9 unique scenes featuring 5 object instances. For each object and its respective tasks, we conduct two trials per framework and place the object back in its original place once grasped. In total, we conduct 774 trials for the KC-USC subsplit and 530 for the UC-USC subsplit. We measure the accuracy of each module in terms of success rates, which is the number of successes over the number of attempts. In dataset experiments, success can be partial (i.e. be 0.5 instead of 1) in cases where the task succeeded but the prediction was not ideal (e.g. the object segmentation only segmenting half of the object but managing to predict and execute the correct task-oriented grasp).

\subsection{Experiments on TD-TOG}
The following section summarizes the performance of each framework on the TD-TOG dataset in standard task-oriented grasping and differentiating between object subcategories.

\subsubsection{Overall Performance and Generalization}
The S$^2$A-Net model achieves an average per-object grasp accuracy of 88.9\%, 83.1\%, and 80.0\% for the KC-KSC, KC-USC, and UC-USC subsplits, respectively. Thus, S$^2$A-Net achieves an overall grasp accuracy of 84.0\% across all subsplits of the \textit{category} split, demonstrating the dataset's effectiveness in training the model to predict grasps for both new scenes and object categories. Table \ref{tab:tdtog_object_segmentation_results.tex} shows that the addition of the SSF module significantly improves the performance of the SAM model, generating an average percentage increase of 192.8\% in AP$_{50}$, and 148.2\% in IoU score. This highlights the effectiveness of the SSF module in removing noise and distracting segment masks.

\begin{table}
\caption{Object segmentation results on the TD-TOG dataset for the category split}
\resizebox{\linewidth}{!}{%
\begin{tabular}{lcccccccc}
\hline
          &        &       & \multicolumn{6}{c}{Mask AP Scores}   \\ \cline{4-9} 
\multirow{-2}{*}{Model} &
  \multirow{-2}{*}{Subsplit} &
  \multirow{-2}{*}{\begin{tabular}[c]{@{}c@{}}IoU\\ Score\end{tabular}} &
  AP &
  AP$_{50}$ &
  AP$_{75}$ &
  AP$_{\text{s}}$ &
  AP$_{\text{m}}$ &
  AP$_{\text{l}}$ \\ \hline
SAM       & KC-KSC & 0.343 & 0.244 & 0.337 & 0.245 & 0.057 & 0.496 & 0.137 \\
SAM + SSF & KC-KSC & 0.867 & 0.764 & 0.960 & 0.796 & 0.529 & 0.822 & 0.878 \\
\rowcolor[HTML]{e7e6e8} 
SAM       & KC-USC & 0.320 & 0.255 & 0.324 & 0.296 & 0.177 & 0.447 & 0.140 \\
\rowcolor[HTML]{e7e6e8} 
SAM + SSF & KC-USC & 0.806 & 0.709 & 0.869 & 0.831 & 0.763 & 0.698 & 0.972 \\
SAM       & UC-USC & 0.320 & 0.216 & 0.268 & 0.240 & 0.054 & 0.376 & 0.154 \\
SAM + SSF & UC-USC & 0.768 & 0.710 & 0.872 & 0.800 & 0.748 & 0.691 & 0.988 \\ \hline
\end{tabular}%
}
\label{tab:tdtog_object_segmentation_results.tex}
\end{table}
\begin{table}
\caption{Object recognition, affordance recognition and TOG detection results on the TD-TOG dataset for single-object scenes of the category split}
\resizebox{\linewidth}{!}{%
\begin{tabular}{clccc}
\multicolumn{1}{l}{} &
   &
  \begin{tabular}[c]{@{}c@{}}Object \\ Recognition\end{tabular} &
  \begin{tabular}[c]{@{}c@{}}Affordance\\ Recognition\end{tabular} &
  \begin{tabular}[c]{@{}c@{}}Task-oriented \\ Grasp Detection\end{tabular} \\ \hline
\multicolumn{1}{l}{Subsplit}                     & Framework    & F$_1$ Score & F$_1$ Score & TOG Accuracy \\ \hline
                                                 & Binary-TOG   & 0.875       & 0.795       & 90.9        \\
                                                 & OS-TOG       & 0.875       & 0.795       & 88.6        \\
\multirow{-3}{*}{KC-KSC}                         & Standard-TOG & 0.875       & 0.852       & 90.9        \\
\rowcolor[HTML]{e7e6e8} 
\cellcolor[HTML]{e7e6e8}                         & Binary-TOG   & 1.000       & 0.757       & 81.8        \\
\rowcolor[HTML]{e7e6e8} 
\cellcolor[HTML]{e7e6e8}                         & OS-TOG       & 0.875       & 0.735       & 77.3        \\
\rowcolor[HTML]{e7e6e8} 
\multirow{-3}{*}{\cellcolor[HTML]{e7e6e8}KC-USC} & Standard-TOG & 0.875       & 0.727       & 65.9        \\
                                                 & Binary-TOG   & 1.000       & 0.761       & 69.8        \\
                         & OS-TOG       & 1.000       & 0.766       & 67.4        \\ 

\multirow{-2}{*}{UC-USC} & Standard-TOG       & --    & --    & --        \\\hline
\end{tabular}%
}
\label{tab:tdtog_overall_singleobject_results}
\end{table}
\begin{table}
\caption{Object recognition, affordance recognition and TOG detection results on the TD-TOG dataset for multi-object scenes of the category split}
\resizebox{\linewidth}{!}{%
\begin{tabular}{clccc}
\multicolumn{1}{l}{} &
   &
  \begin{tabular}[c]{@{}c@{}}Object \\ Recognition\end{tabular} &
  \begin{tabular}[c]{@{}c@{}}Affordance\\ Recognition\end{tabular} &
  \begin{tabular}[c]{@{}c@{}}Task-oriented \\ Grasp Detection\end{tabular} \\ \hline
\multicolumn{1}{l}{Subsplit}                        & Framework    & F$_1$ Score & F$_1$ Score & TOG Accuracy \\ \hline
                                                 & Binary-TOG   & 0.893    & 0.768    & 82.6        \\
                                                 & OS-TOG       & 0.965    & 0.782    & 83.9        \\
\multirow{-3}{*}{KC-KSC}                         & Standard-TOG & 0.907    & 0.853    & 90.6        \\
\rowcolor[HTML]{e7e6e8} 
\cellcolor[HTML]{e7e6e8}                         & Binary-TOG   & 0.873    & 0.699    & 69.5        \\
\rowcolor[HTML]{e7e6e8} 
\cellcolor[HTML]{e7e6e8}                         & OS-TOG       & 0.886    & 0.714    & 70.8        \\
\rowcolor[HTML]{e7e6e8} 
\multirow{-3}{*}{\cellcolor[HTML]{e7e6e8}KC-USC} & Standard-TOG & 0.715    & 0.542    & 52.3        \\
                                                 & Binary-TOG   & 0.733    & 0.610    & 54.5        \\
                                                & OS-TOG       & 0.938    & 0.719    & 63.3        \\
\multirow{-3}{*}{UC-USC}                         & Standard-TOG       & --    & --    & --        \\\hline
\end{tabular}%
}
\label{tab:tdtog_overall_multiobject_results}
\end{table}
\begin{table}[!hbtp]
\caption{Task-oriented grasp (TG) accuracy on TD-TOG for the category split and KC-USC subsplit}
\resizebox{0.95\linewidth}{!}{%
\begin{tabular}{clccc}
\hline
\textbf{Object}                                      & \textbf{Task}      & Binary-TOG & OS-TOG & Standard-TOG \\ \hline
\rowcolor[HTML]{EFEFEF} 
\cellcolor[HTML]{EFEFEF}                             & \textit{transport} & 86.7       & 93.3   & 86.7         \\
\rowcolor[HTML]{EFEFEF} 
\cellcolor[HTML]{EFEFEF}                             & \textit{handover}  & 86.7       & 86.7   & 0.0          \\
\rowcolor[HTML]{EFEFEF} 
\multirow{-3}{*}{\cellcolor[HTML]{EFEFEF}spoon}      & \textit{scooping}  & 86.7       & 86.7   & 20.0         \\
                                                     & \textit{transport} & 100.0      & 100.0  & 100.0        \\
                                                     & \textit{handover}  & 60.0       & 53.3   & 20.0         \\
\multirow{-3}{*}{pen}                                & \textit{writing}   & 86.7       & 86.7   & 80.0         \\
\rowcolor[HTML]{EFEFEF} 
\cellcolor[HTML]{EFEFEF}                             & \textit{transport} & 93.3       & 93.3   & 86.7         \\
\rowcolor[HTML]{EFEFEF} 
\cellcolor[HTML]{EFEFEF}                             & \textit{handover}  & 60.0       & 60.0   & 86.7         \\
\rowcolor[HTML]{EFEFEF} 
\multirow{-3}{*}{\cellcolor[HTML]{EFEFEF}hammer}     & \textit{hitting}   & 60.0       & 73.3   & 86.7         \\
                                                     & \textit{transport} & 80.0       & 80.0   & 66.7         \\
                                                     & \textit{handover}  & 80.0       & 80.0   & 46.7         \\
\multirow{-3}{*}{clip}                               & \textit{clamping}  & 0.0        & 0.0    & 0.0          \\
\rowcolor[HTML]{EFEFEF} 
\cellcolor[HTML]{EFEFEF}                             & \textit{transport} & 100.0      & 100.0  & 100.0        \\
\rowcolor[HTML]{EFEFEF} 
\multirow{-2}{*}{\cellcolor[HTML]{EFEFEF}controller} & \textit{handover}  & 86.7       & 86.7   & 20.0         \\
                                                     & \textit{transport} & 93.3       & 100.0  & 13.3         \\
                                                     & \textit{handover}  & 73.3       & 80.0   & 13.3         \\
\multirow{-3}{*}{razor}                              & \textit{shaving}   & 86.7       & 73.3   & 13.3         \\
\rowcolor[HTML]{EFEFEF} 
\cellcolor[HTML]{EFEFEF}                             & \textit{transport} & 86.7       & 100.0  & 93.3         \\
\rowcolor[HTML]{EFEFEF} 
\cellcolor[HTML]{EFEFEF}                             & \textit{handover}  & 86.7       & 100.0  & 0.0          \\
\rowcolor[HTML]{EFEFEF} 
\multirow{-3}{*}{\cellcolor[HTML]{EFEFEF}grater}     & \textit{grating}   & 40.0       & 46.7   & 33.3         \\
                                                     & \textit{transport} & 53.3       & 40.0   & 40.0         \\
                                                     & \textit{handover}  & 40.0       & 26.7   & 13.3         \\
\multirow{-3}{*}{scissors}                           & \textit{cutting}   & 26.7       & 33.3   & 33.3         \\
\rowcolor[HTML]{EFEFEF} 
\cellcolor[HTML]{EFEFEF}                             & \textit{transport} & 60.0       & 73.3   & 86.7         \\
\rowcolor[HTML]{EFEFEF} 
\cellcolor[HTML]{EFEFEF}                             & \textit{handover}  & 20.0       & 13.3   & 26.7         \\
\rowcolor[HTML]{EFEFEF} 
\multirow{-3}{*}{\cellcolor[HTML]{EFEFEF}dish\_brush} & \textit{scrubbing} & 33.3 & 33.3 & 73.3 \\
                                                     & \textit{transport} & 100.0      & 86.7   & 46.7         \\
                                                     & \textit{handover}  & 60.0       & 40.0   & 13.3         \\
\multirow{-3}{*}{paint\_brush}                       & \textit{painting}  & 100.0      & 86.7   & 0.0          \\
\rowcolor[HTML]{EFEFEF} 
\cellcolor[HTML]{EFEFEF}                             & \textit{transport} & 93.3       & 86.7   & 93.3         \\
\rowcolor[HTML]{EFEFEF} 
\cellcolor[HTML]{EFEFEF}                             & \textit{handover}  & 40.0       & 33.3   & 93.3         \\
\rowcolor[HTML]{EFEFEF} 
\multirow{-3}{*}{\cellcolor[HTML]{EFEFEF}toothpaste} & \textit{opening}   & 40.0       & 26.7   & 100.0        \\
                                                     & \textit{transport} & 60.0       & 86.7   & 86.7         \\
                                                     & \textit{handover}  & 40.0       & 46.7   & 46.7         \\
\multirow{-3}{*}{glue}                               & \textit{gluing}    & 46.7       & 80.0   & 66.7         \\
\rowcolor[HTML]{EFEFEF} 
\cellcolor[HTML]{EFEFEF}                             & \textit{transport} & 86.7       & 86.7   & 86.7         \\
\rowcolor[HTML]{EFEFEF} 
\cellcolor[HTML]{EFEFEF}                             & \textit{handover}  & 80.0       & 80.0   & 93.3         \\
\rowcolor[HTML]{EFEFEF} 
\multirow{-3}{*}{\cellcolor[HTML]{EFEFEF}pan}        & \textit{frying}    & 86.7       & 86.7   & 86.7         \\
                                                     & \textit{transport} & 80.0       & 80.0   & 66.7         \\
                                                     & \textit{handover}  & 53.3       & 53.3   & 46.7         \\
\multirow{-3}{*}{screwdriver}                        & \textit{screwing}  & 80.0       & 80.0   & 80.0         \\
\rowcolor[HTML]{EFEFEF} 
\cellcolor[HTML]{EFEFEF}                             & \textit{transport} & 100.0      & 100.0  & 46.7         \\
\rowcolor[HTML]{EFEFEF} 
\cellcolor[HTML]{EFEFEF}                             & \textit{handover}  & 86.7       & 93.3   & 0.0          \\
\rowcolor[HTML]{EFEFEF} 
\multirow{-3}{*}{\cellcolor[HTML]{EFEFEF}spatula}    & \textit{flipping}  & 93.3       & 100.0  & 46.7         \\ \hline
\end{tabular}%
}
\label{tab:tdtog_kcsuc_togacc_by_objtask}
\end{table}
\begin{table}[!hbtp]
\caption{Task-oriented grasp (TG) accuracy on TD-TOG for the category split and UC-USC subsplit}
\centering
\resizebox{\linewidth}{!}{%
\begin{tabular}{clccc}
\hline
\textbf{Object}                                      & \textbf{Task}       & Binary-TOG & OS-TOG & Standard-TOG \\ \hline
\rowcolor[HTML]{EFEFEF} 
\cellcolor[HTML]{EFEFEF}                             & \textit{transport}  & 86.7       & 93.3   & -- \\
\rowcolor[HTML]{EFEFEF} 
\cellcolor[HTML]{EFEFEF}                             & \textit{handover}   & 13.3       & 6.7    & -- \\
\rowcolor[HTML]{EFEFEF} 
\multirow{-3}{*}{\cellcolor[HTML]{EFEFEF}pizza\_cutter} & \textit{cutting}  & 86.7 & 93.3 & -- \\
                                                     & \textit{transport}  & 93.3       & 93.3   & -- \\
                                                     & \textit{handover}   & 93.3       & 93.3   & -- \\
\multirow{-3}{*}{tenderizer}                         & \textit{hitting}    & 60.0       & 60.0   & -- \\
\rowcolor[HTML]{EFEFEF} 
\cellcolor[HTML]{EFEFEF}                             & \textit{transport}  & 93.3       & 86.7   & -- \\
\rowcolor[HTML]{EFEFEF} 
\multirow{-2}{*}{\cellcolor[HTML]{EFEFEF}cable}      & \textit{connecting} & 80.0       & 73.3   & -- \\
                                                     & \textit{transport}  & 40.0       & 46.7   & -- \\
                                                     & \textit{handover}   & 0.0        & 0.0    & -- \\
\multirow{-3}{*}{thermometer}                        & \textit{measuring}  & 33.3       & 66.7   & -- \\
\rowcolor[HTML]{EFEFEF} 
\cellcolor[HTML]{EFEFEF}                             & \textit{transport}  & 86.7       & 100.0  & -- \\
\rowcolor[HTML]{EFEFEF} 
\cellcolor[HTML]{EFEFEF}                             & \textit{handover}   & 53.3       & 60.0   & -- \\
\rowcolor[HTML]{EFEFEF} 
\multirow{-3}{*}{\cellcolor[HTML]{EFEFEF}marker}     & \textit{writing}    & 73.3       & 80.0   & -- \\
                                                     & \textit{transport}  & 20.0       & 13.3   & -- \\
\multirow{-2}{*}{plush\_toy}                         & \textit{handover}   & 6.7        & 0.0    & -- \\
\rowcolor[HTML]{EFEFEF} 
\cellcolor[HTML]{EFEFEF}                             & \textit{transport}  & 66.7       & 93.3   & -- \\
\rowcolor[HTML]{EFEFEF} 
\cellcolor[HTML]{EFEFEF}                             & \textit{handover}   & 33.3       & 46.7   & -- \\
\rowcolor[HTML]{EFEFEF} 
\multirow{-3}{*}{\cellcolor[HTML]{EFEFEF}tongs}      & \textit{clamping}   & 40.0       & 66.7   & -- \\
                                                     & \textit{transport}  & 86.7       & 86.7   & -- \\
                                                     & \textit{handover}   & 0.0        & 0.0    & -- \\
\multirow{-3}{*}{measuring\_cup}                     & \textit{scooping}   & 100.0      & 100.0  & -- \\
\rowcolor[HTML]{EFEFEF} 
\cellcolor[HTML]{EFEFEF}                             & \textit{transport}  & 100.0      & 100.0  & -- \\
\rowcolor[HTML]{EFEFEF} 
\cellcolor[HTML]{EFEFEF}                             & \textit{handover}   & 26.7       & 26.7   & -- \\
\rowcolor[HTML]{EFEFEF} 
\multirow{-3}{*}{\cellcolor[HTML]{EFEFEF}toothbrush} & \textit{brushing}   & 73.3       & 66.7   & -- \\
                                                     & \textit{transport}  & 93.3       & 93.3   & -- \\
                                                     & \textit{handover}   & 40.0       & 40.0   & -- \\
\multirow{-3}{*}{dustpan}                            & \textit{sweeping}   & 93.3       & 93.3   & -- \\
\rowcolor[HTML]{EFEFEF} 
\cellcolor[HTML]{EFEFEF}                             & \textit{transport}  & 50.0       & 78.6   & -- \\
\rowcolor[HTML]{EFEFEF} 
\cellcolor[HTML]{EFEFEF}                             & \textit{handover}   & 28.6       & 50.0   & -- \\
\rowcolor[HTML]{EFEFEF} 
\multirow{-3}{*}{\cellcolor[HTML]{EFEFEF}paint\_roller} & \textit{painting} & 28.6 & 57.1 & -- \\
                                                     & \textit{transport}  & 40.0       & 100.0  & -- \\
                                                     & \textit{handover}   & 26.7       & 53.3   & -- \\
\multirow{-3}{*}{screw}                              & \textit{screwing}   & 33.3       & 60.0   & -- \\
\rowcolor[HTML]{EFEFEF} 
\cellcolor[HTML]{EFEFEF}                             & \textit{transport}  & 93.3       & 100.0  & -- \\
\rowcolor[HTML]{EFEFEF} 
\cellcolor[HTML]{EFEFEF}                             & \textit{handover}   & 40.0       & 33.3   & -- \\
\rowcolor[HTML]{EFEFEF} 
\multirow{-3}{*}{\cellcolor[HTML]{EFEFEF}vitamin}    & \textit{opening}    & 20.0       & 20.0   & -- \\
                                                     & \textit{transport}  & 37.5       & 62.5   & -- \\
                                                     & \textit{handover}   & 12.5       & 25.0   & -- \\
\multirow{-3}{*}{peeler}                             & \textit{peeling}    & 37.5       & 62.5   & -- \\
\rowcolor[HTML]{EFEFEF} 
\cellcolor[HTML]{EFEFEF}                             & \textit{transport}  & 100.0      & 93.3   & -- \\
\rowcolor[HTML]{EFEFEF} 
\cellcolor[HTML]{EFEFEF}                             & \textit{handover}   & 66.7       & 66.7   & -- \\
\rowcolor[HTML]{EFEFEF} 
\multirow{-3}{*}{\cellcolor[HTML]{EFEFEF}hairbrush}  & \textit{brushing}   & 100.0      & 93.3   & -- \\ \hline
\end{tabular}%
}
\label{tab:tdtog_ucsuc_togacc_by_objtask}
\end{table}

Tables \ref{tab:tdtog_overall_singleobject_results} and \ref{tab:tdtog_overall_multiobject_results} depict the framework's overall performance for object recognition, affordance recognition, and task-oriented grasp detection on single-object and multi-object scenes, respectively. Note that each module's performance is dependent on the preceding module. For instance, the object recognition module relies on the accuracy of the object segmentation module and, hence, is affected by failures to segment the target object or partial segmentations. It is also essential to highlight that the object recognition model used in Binary-TOG was not directly trained on TD-TOG. As it uses CLIP, a foundation model trained on large-scale image-text pairs from diverse domains, it may not have encountered certain object categories or subcategories present in the TD-TOG dataset. Consequently, the framework is likely to have increased difficulty in object recognition across all subsplits, particularly for objects absent from its pretraining data. On the other hand, the models of OS-TOG and Standard-TOG were trained on objects found in the KC-KSC subsplit.

When comparing the task-oriented grasp (TOG) accuracy of Standard-TOG, OS-TOG and Binary-TOG across multi-object and single-object scenes for the KC-KSC and KC-USC subsplits (Tables \ref{tab:tdtog_overall_singleobject_results} and \ref{tab:tdtog_overall_multiobject_results}), Standard-TOG has the largest drop in performance, emphasizing its limited ability to generalize to new subcategories. In contrast, OS-TOG and Binary-TOG maintain relatively high TOG accuracy, demonstrating the generalization capabilities of one-shot and zero-shot techniques. Furthermore, Standard-TOG is unable to participate in the UC-USC subsplit without retraining, which further emphasizes the adaptability of the latter two frameworks, especially towards novel object categories. Nonetheless, the results show that training the sub-models of Standard-TOG on TD-TOG enabled it to achieve an average TOG accuracy of 90.8\% for known objects but new scenes, and 59.1\% for novel object subcategories across single-object and multi-object scenes. 

Results in Table \ref{tab:tdtog_overall_singleobject_results} indicate that Binary-TOG outperforms OS-TOG and achieves the highest TOG accuracy in single-object scenes across all subsplits. However, in multi-object scenes, OS-TOG demonstrates better performance across all subsplits, suggesting that it is more robust to the presence of distractor objects (Table \ref{tab:tdtog_overall_multiobject_results}). This performance difference can be attributed to the object recognition module, which is the primary distinction between the two frameworks. 

Table \ref{tab:tdtog_kcsuc_togacc_by_objtask} highlights that Binary-TOG mostly performed worse than OS-TOG in the KC-USC subsplit when identifying the \texttt{glue}, \texttt{grater} and \texttt{spatula} which have different shapes than other instances of the respective categories. This adds a consideration to take into account when using zero-shot approaches, that they may be sensitive to objects that deviate from their ordinary forms. This is due to the fact that they are often trained on data from online sources, which can lead to biases in how certain objects are visually represented. This is also the case in the UC-USC subsplit (Table \ref{tab:tdtog_ucsuc_togacc_by_objtask}), where Binary-TOG performs worse than OS-TOG on the \texttt{paint roller} and \texttt{peeler}.

\subsubsection{Differentiating between Subcategories}
Table \ref{tab:tdtog_overall_kc_results} shows that Binary-TOG scores an average F$_1$ score of 0.714 and OS-TOG scores 0.806 in differentiating between object subcategories in the KC subsplit. This observed performance difference is expected as OS-TOG has an advantage in that it has encountered most of the objects when training the object recognition module. However, for novel object categories (Table \ref{tab:tdtog_overall_uc_results}), OS-TOG still outperforms Binary-TOG in object recognition with an average F$_1$ score of 0.819 as opposed to a 0.682. This signifies that OS-TOG is overall superior to Binary-TOG in differentiating between object subcategories and does not overfit to instances it encountered in training. 

\begin{table}
\caption{Object recognition results of Binary-TOG and OS-TOG on the TD-TOG subcategory split and KC subsplit}
\centering
\resizebox{0.9\linewidth}{!}{%
\begin{tabular}{lcccc}
                     & \multicolumn{2}{c}{Binary-TOG} & \multicolumn{2}{c}{OS-TOG} \\ \cline{2-5} 
\textbf{Category} & Accuracy       & F$_1$ Score      & Accuracy        & F$_1$ Score       \\ \hline
glue                 & 0.528          & 0.431         & 0.917           & 0.914          \\
pen                  & 0.694          & 0.680         & 0.667           & 0.650          \\
screwdriver          & 0.750          & 0.719         & 0.833           & 0.830          \\
razor                & 0.694          & 0.668         & 0.972           & 0.972          \\
spatula              & 0.750          & 0.754         & 0.861           & 0.861          \\
clip                 & 0.861          & 0.857         & 0.806           & 0.782          \\
dish\_brush          & 0.667          & 0.660         & 0.972           & 0.972          \\
controller           & 0.861          & 0.843         & 1.000           & 1.000          \\
grater               & 0.667          & 0.646         & 0.778           & 0.782          \\
toothpaste           & 0.917          & 0.917         & 0.583           & 0.512          \\
hammer               & 1.000          & 1.000         & 0.750           & 0.729          \\
scissors             & 0.806          & 0.801         & 0.778           & 0.795          \\
paint\_brush         & 0.528          & 0.489         & 0.750           & 0.754          \\
pan                  & 0.528          & 0.526         & 0.750           & 0.746          \\
spoon                & 0.750          & 0.718         & 0.806           & 0.797 
\\ \hline
\end{tabular}% 
}
\label{tab:tdtog_overall_kc_results}
\end{table}
\begin{table}
\caption{Object recognition results of Binary-TOG and OS-TOG on the TD-TOG subcategory split and UC subsplit}
\centering
\resizebox{0.95\linewidth}{!}{%
\begin{tabular}{lcccc}
                  & \multicolumn{2}{c}{Binary-TOG} & \multicolumn{2}{c}{OS-TOG} \\ \cline{2-5} 
\textbf{Category} & Accuracy       & F$_1$ Score      & Accuracy     & F$_1$ Score    \\ \hline
toothbrush        & 0.861          & 0.849         & 0.944        & 0.942       \\
thermometer       & 0.778          & 0.774         & 0.917        & 0.916       \\
vitamin           & 0.583          & 0.551         & 0.861        & 0.863       \\
measuring\_cup    & 0.611          & 0.589         & 0.889        & 0.887       \\
hairbrush         & 0.750          & 0.654         & 1.000        & 1.000       \\
dustpan           & 0.917          & 0.914         & 0.833        & 0.831       \\
tenderizer        & 0.722          & 0.708         & 0.806        & 0.771       \\
peeler            & 0.833          & 0.833         & 0.750        & 0.743       \\
paint\_roller     & 0.500          & 0.466         & 0.778        & 0.775       \\
pizza\_cutter     & 0.611          & 0.612         & 0.833        & 0.833       \\
screw             & 0.417          & 0.360         & 0.611        & 0.617       \\
cable             & 0.722          & 0.718         & 0.917        & 0.912       \\
plush\_toy        & 0.722          & 0.758         & 0.861        & 0.893       \\
tongs             & 0.806          & 0.780         & 0.667        & 0.657       \\
marker            & 0.694          & 0.661         & 0.667        & 0.639       \\ \hline
\end{tabular}%
}
\label{tab:tdtog_overall_uc_results}
\end{table}
\begin{table*}
\caption{The overall success rates of the frameworks in real-world experiments for different subsplits of objects from the TD-TOG category split. For instance, UC-USC means that real-world experiments were carried out on the same set of objects that were in the UC-USC subsplit}
\resizebox{\textwidth}{!}{%
\begin{tabular}{llccccccccc}
\textbf{Subsplit} &
  \textbf{Framework} &
  \begin{tabular}[c]{@{}c@{}}Obj. Seg.\\ Rate (\%)\end{tabular} &
  \begin{tabular}[c]{@{}c@{}}Obj. Rec.\\ Rate (\%)\end{tabular} &
  \begin{tabular}[c]{@{}c@{}}Aff. Rec.\\ Rate (\%)\end{tabular} &
  \begin{tabular}[c]{@{}c@{}}Grasp Det.\\ Rate (\%)\end{tabular} &
  \begin{tabular}[c]{@{}c@{}}TOG Det.\\ Rate (\%)\end{tabular} &
  \begin{tabular}[c]{@{}c@{}}Grasp Succ.\\ Rate (\%)\end{tabular} &
  \begin{tabular}[c]{@{}c@{}}TOG Succ.\\ Rate (\%)\end{tabular} &
  \begin{tabular}[c]{@{}c@{}}Avg. Inference\\ Time (in sec)\end{tabular} &
  \begin{tabular}[c]{@{}c@{}}Avg. Grasp\\ Time (in sec)\end{tabular} \\ \hline
KC-USC & Binary-TOG   & 92.6 & 94.5 & 75.2 & 91.3 & 79.5 & 76.7 & 64.8 & 8.4 & 8.9 \\
KC-USC & OS-TOG       & 92.4 & 93.8 & 76.5 & 93.6 & 80.9 & 76.3 & 65.3 & 8.8 & 9.0 \\
KC-USC & Standard-TOG & 91.5 & 80.9 & 54.7 & 77.3 & 57.0 & 57.2 & 44.1 & 5.0 & 8.8 \\
\rowcolor[HTML]{EFEFEF} 
UC-USC & Binary-TOG   & 93.4 & 70.2 & 62.6 & 84.9 & 62.8 & 68.6 & 52.7 & 9.3 & 8.9 \\
\rowcolor[HTML]{EFEFEF} 
UC-USC & OS-TOG       & 93.6 & 90.3 & 77.5 & 93.0 & 79.8 & 81.6 & 70.9 & 9.8 & 8.8 \\ \hline
\end{tabular}%
}
\label{tab:world_overall_results}
\end{table*}
\begin{figure*}
    \centering
    \includegraphics[width=\linewidth]{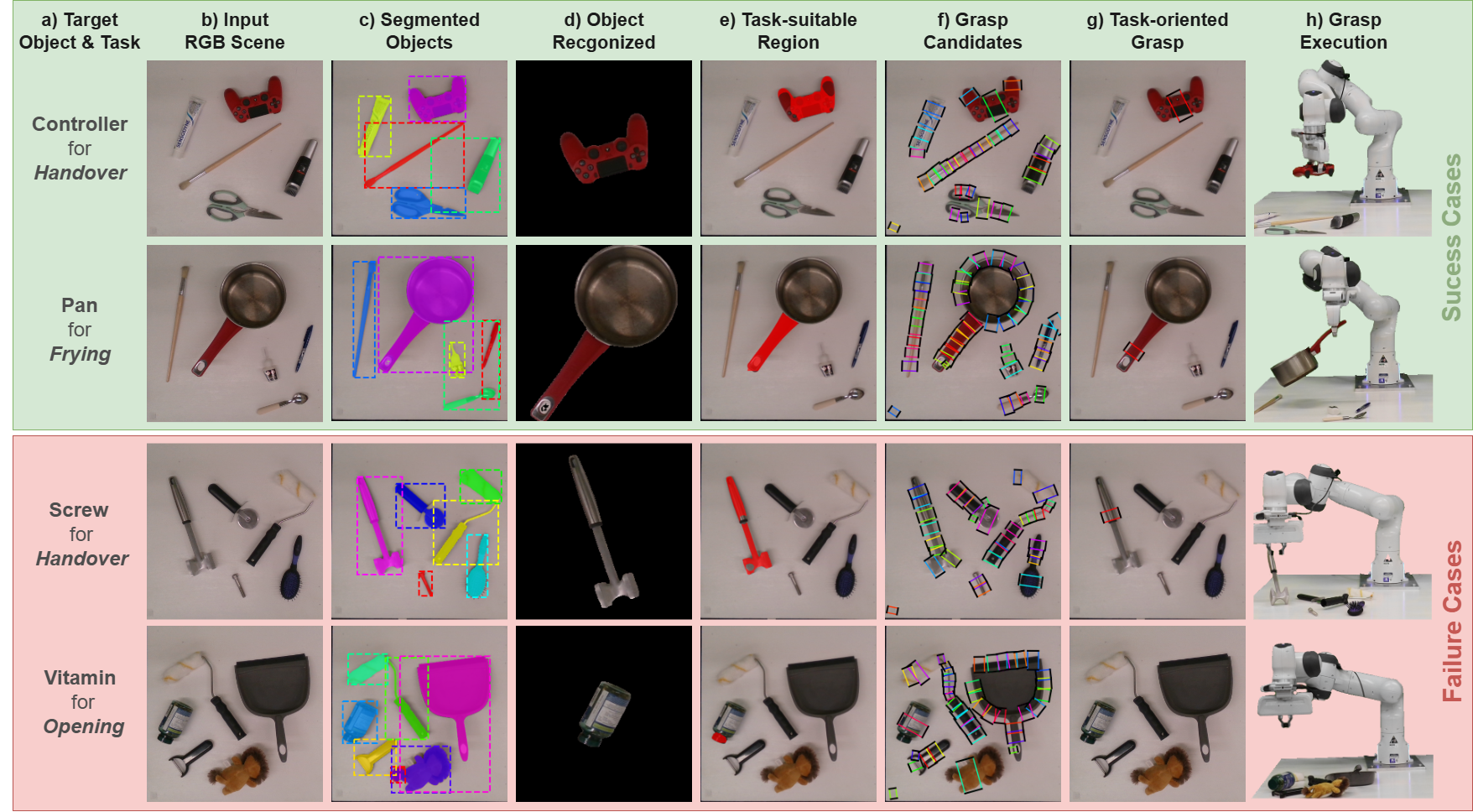}
    \caption{Example success and failure cases of Binary-TOG in real-world experiments. For each case, the predictions given by each module in the framework for an input target object and target task \textbf{(a)} and input RGB scene \textbf{(b)} are shown. This includes the objects segmented by the zero-shot object segmentation module \textbf{(c)}, the object identified from the segments as the target object \textbf{(d)}, the task-suitable region identified in the one-shot affordance recognition module \textbf{(e)} and the grasp candidates \textbf{(f)}, the final task-oriented grasp determined by the zero-shot task-oriented grasp module \textbf{(g)}, and the execution of the predicted grasp \textbf{(h)}. }
    \label{fig:real_world_results}
\end{figure*}
\begin{table}
\caption{The TOG success rate of each framework in real-world experiments on objects and tasks from the KC-USC subsplit of the TD-TOG dataset}
\resizebox{0.95\linewidth}{!}{%
\begin{tabular}{llccc}
\textbf{Object}                                       & \textbf{Task}      & OS-TOG & Binary-TOG & Standard-TOG \\ \hline
\rowcolor[HTML]{EFEFEF} 
\cellcolor[HTML]{EFEFEF}                              & \textit{transport} & 75.0   & 33.3       & 25.0         \\
\rowcolor[HTML]{EFEFEF} 
\multirow{-2}{*}{\cellcolor[HTML]{EFEFEF}controller} & \textit{handover} & 100.0 & 100.0 & 16.7 \\
                                                      & \textit{transport} & 83.3   & 100.0      & 50.0         \\
                                                      & \textit{handover}  & 83.3   & 100.0      & 33.3         \\
\multirow{-3}{*}{paint\_brush}                        & \textit{painting}  & 100.0  & 100.0      & 0.0          \\
\rowcolor[HTML]{EFEFEF} 
\cellcolor[HTML]{EFEFEF}                              & \textit{transport} & 91.7   & 66.7       & 0.0          \\
\rowcolor[HTML]{EFEFEF} 
\cellcolor[HTML]{EFEFEF}                              & \textit{handover}  & 25.0   & 58.3       & 25.0         \\
\rowcolor[HTML]{EFEFEF} 
\multirow{-3}{*}{\cellcolor[HTML]{EFEFEF}razor}       & \textit{shaving}   & 100.0  & 83.3       & 0.0          \\
                                                      & \textit{transport} & 41.7   & 33.3       & 33.3         \\
                                                      & \textit{handover}  & 16.7   & 33.3       & 0.0          \\
\multirow{-3}{*}{scissors}                            & \textit{cutting}   & 33.3   & 16.7       & 0.0          \\
\rowcolor[HTML]{EFEFEF} 
\cellcolor[HTML]{EFEFEF}                              & \textit{transport} & 66.7   & 75.0       & 100.0        \\
\rowcolor[HTML]{EFEFEF} 
\cellcolor[HTML]{EFEFEF}                              & \textit{handover}  & 58.3   & 83.3       & 100.0        \\
\rowcolor[HTML]{EFEFEF} 
\multirow{-3}{*}{\cellcolor[HTML]{EFEFEF}toothpaste}  & \textit{opening}   & 50.0   & 50.0       & 83.3         \\
                                                      & \textit{transport} & 100.0  & 100.0      & 100.0        \\
                                                      & \textit{handover}  & 100.0  & 100.0      & 66.7         \\
\multirow{-3}{*}{clip}                                & \textit{clamping}  & 0.0    & 0.0        & 0.0          \\
\rowcolor[HTML]{EFEFEF} 
\cellcolor[HTML]{EFEFEF}                              & \textit{transport} & 100.0  & 100.0      & 100.0        \\
\rowcolor[HTML]{EFEFEF} 
\cellcolor[HTML]{EFEFEF}                              & \textit{handover}  & 100.0  & 100.0      & 100.0        \\
\rowcolor[HTML]{EFEFEF} 
\multirow{-3}{*}{\cellcolor[HTML]{EFEFEF}screwdriver} & \textit{screwing}  & 83.3   & 66.7       & 50.0         \\
                                                      & \textit{transport} & 83.3   & 83.3       & 50.0         \\
                                                      & \textit{handover}  & 66.7   & 66.7       & 50.0         \\
\multirow{-3}{*}{spatula}                             & \textit{flipping}  & 66.7   & 66.7       & 33.3         \\
\rowcolor[HTML]{EFEFEF} 
\cellcolor[HTML]{EFEFEF}                              & \textit{transport} & 16.7   & 33.3       & 83.3         \\
\rowcolor[HTML]{EFEFEF} 
\cellcolor[HTML]{EFEFEF}                              & \textit{handover}  & 0.0    & 33.3       & 0.0          \\
\rowcolor[HTML]{EFEFEF} 
\multirow{-3}{*}{\cellcolor[HTML]{EFEFEF}dish\_brush} & \textit{scrubbing} & 33.3   & 0.0        & 33.3         \\
                                                      & \textit{transport} & 0.0    & 50.0       & 50.0         \\
                                                      & \textit{handover}  & 83.3   & 83.3       & 50.0         \\
\multirow{-3}{*}{pan}                                 & \textit{frying}    & 66.7   & 33.3       & 50.0         \\
\rowcolor[HTML]{EFEFEF} 
\cellcolor[HTML]{EFEFEF}                              & \textit{transport} & 100.0  & 100.0      & 100.0        \\
\rowcolor[HTML]{EFEFEF} 
\cellcolor[HTML]{EFEFEF}                              & \textit{handover}  & 100.0  & 66.7       & 0.0          \\
\rowcolor[HTML]{EFEFEF} 
\multirow{-3}{*}{\cellcolor[HTML]{EFEFEF}spoon}       & \textit{scooping}  & 100.0  & 100.0      & 0.0          \\
                                                      & \textit{transport} & 41.7   & 58.3       & 50.0         \\
                                                      & \textit{handover}  & 16.7   & 41.7       & 0.0          \\
\multirow{-3}{*}{grater}                              & \textit{grating}   & 0.0    & 0.0        & 0.0          \\
\rowcolor[HTML]{EFEFEF} 
\cellcolor[HTML]{EFEFEF}                              & \textit{transport} & 66.7   & 58.3       & 66.7         \\
\rowcolor[HTML]{EFEFEF} 
\cellcolor[HTML]{EFEFEF}                              & \textit{handover}  & 83.3   & 83.3       & 50.0         \\
\rowcolor[HTML]{EFEFEF} 
\multirow{-3}{*}{\cellcolor[HTML]{EFEFEF}glue}        & \textit{gluing}    & 8.3    & 0.0        & 0.0          \\
                                                      & \textit{transport} & 100.0  & 100.0      & 83.3         \\
                                                      & \textit{handover}  & 75.0   & 66.7       & 25.0         \\
\multirow{-3}{*}{pen}                                 & \textit{writing}   & 100.0  & 83.3       & 58.3         \\
\rowcolor[HTML]{EFEFEF} 
\cellcolor[HTML]{EFEFEF}                              & \textit{transport} & 83.3   & 100.0      & 75.0         \\
\rowcolor[HTML]{EFEFEF} 
\cellcolor[HTML]{EFEFEF}                              & \textit{handover}  & 100.0  & 100.0      & 100.0        \\
\rowcolor[HTML]{EFEFEF} 
\multirow{-3}{*}{\cellcolor[HTML]{EFEFEF}hammer}      & hitting            & 75.0   & 41.7       & 50.0         \\ \hline
\end{tabular}%
}
\label{tab:world_togsuccess_kc_results}
\end{table}
\begin{table}
\caption{The TOG success rate of each framework in real-world experiments on objects and tasks from the UC-USC subsplit of the TD-TOG dataset}
\centering
\resizebox{\linewidth}{!}{%
\begin{tabular}{llccc}
\textbf{Object}                                  & \textbf{Task}       & OS-TOG & Hybrid-TOG & Standard-TOG \\ \hline
\rowcolor[HTML]{EFEFEF} 
\cellcolor[HTML]{EFEFEF}                         & \textit{transport}  & 100.0  & 91.7       & -- \\
\rowcolor[HTML]{EFEFEF} 
\cellcolor[HTML]{EFEFEF}                         & \textit{handover}   & 33.3   & 33.3       & -- \\
\rowcolor[HTML]{EFEFEF} 
\multirow{-3}{*}{\cellcolor[HTML]{EFEFEF}dustpan}       & \textit{sweeping}  & 100.0 & 100.0 & -- \\
                                                 & \textit{transport}  & 100.0  & 100.0      & -- \\
                                                 & \textit{handover}   & 100.0  & 83.3       & -- \\
\multirow{-3}{*}{hairbrush}                      & \textit{brushing}   & 100.0  & 100.0      & -- \\
\rowcolor[HTML]{EFEFEF} 
\cellcolor[HTML]{EFEFEF}                         & \textit{transport}  & 100.0  & 100.0      & -- \\
\rowcolor[HTML]{EFEFEF} 
\cellcolor[HTML]{EFEFEF}                         & \textit{handover}   & 100.0  & 100.0      & -- \\
\rowcolor[HTML]{EFEFEF} 
\multirow{-3}{*}{\cellcolor[HTML]{EFEFEF}marker}        & \textit{writing}   & 83.3  & 100.0 & -- \\
                                                 & \textit{transport}  & 50.0   & 33.3       & -- \\
\multirow{-2}{*}{plush\_toy}                     & \textit{handover}   & 33.3   & 50.0       & -- \\
\rowcolor[HTML]{EFEFEF} 
\cellcolor[HTML]{EFEFEF}                         & \textit{transport}  & 66.7   & 33.3       & -- \\
\rowcolor[HTML]{EFEFEF} 
\cellcolor[HTML]{EFEFEF}                         & \textit{handover}   & 33.3   & 0.0        & -- \\
\rowcolor[HTML]{EFEFEF} 
\multirow{-3}{*}{\cellcolor[HTML]{EFEFEF}thermometer}   & \textit{measuring} & 66.7  & 50.0  & -- \\
                                                 & \textit{transport}  & 66.7   & 100.0      & -- \\
                                                 & \textit{handover}   & 83.3   & 83.3       & -- \\
\multirow{-3}{*}{measuring\_cup}                 & \textit{scooping}   & 83.3   & 83.3       & -- \\
\rowcolor[HTML]{EFEFEF} 
\cellcolor[HTML]{EFEFEF}                         & \textit{transport}  & 66.7   & 33.3       & -- \\
\rowcolor[HTML]{EFEFEF} 
\cellcolor[HTML]{EFEFEF}                         & \textit{handover}   & 100.0  & 66.7       & -- \\
\rowcolor[HTML]{EFEFEF} 
\multirow{-3}{*}{\cellcolor[HTML]{EFEFEF}tenderizer}    & \textit{hitting}   & 33.3  & 0.0   & -- \\
                                                 & \textit{transport}  & 100.0  & 50.0       & -- \\
                                                 & \textit{handover}   & 33.3   & 16.7       & -- \\
\multirow{-3}{*}{tongs}                          & \textit{clamping}   & 83.3   & 33.3       & -- \\
\rowcolor[HTML]{EFEFEF} 
\cellcolor[HTML]{EFEFEF}                         & \textit{transport}  & 100.0  & 100.0      & -- \\
\rowcolor[HTML]{EFEFEF} 
\cellcolor[HTML]{EFEFEF}                         & \textit{handover}   & 16.7   & 41.7       & -- \\
\rowcolor[HTML]{EFEFEF} 
\multirow{-3}{*}{\cellcolor[HTML]{EFEFEF}toothbrush}    & \textit{brushing}  & 100.0 & 83.3  & -- \\
                                                 & \textit{transport}  & 91.7   & 100.0      & -- \\
                                                 & \textit{handover}   & 33.3   & 16.7       & -- \\
\multirow{-3}{*}{vitamin}                        & \textit{opening}    & 16.7   & 33.3       & -- \\
\rowcolor[HTML]{EFEFEF} 
\cellcolor[HTML]{EFEFEF}                         & \textit{transport}  & 66.7   & 16.7       & -- \\
\rowcolor[HTML]{EFEFEF} 
\cellcolor[HTML]{EFEFEF}                         & \textit{handover}   & 66.7   & 16.7       & -- \\
\rowcolor[HTML]{EFEFEF} 
\multirow{-3}{*}{\cellcolor[HTML]{EFEFEF}paint\_roller} & \textit{painting}  & 58.3  & 16.7  & -- \\
                                                 & \textit{transport}  & 100.0  & 100.0      & -- \\
                                                 & \textit{handover}   & 66.7   & 50.0       & -- \\
\multirow{-3}{*}{pizza\_cutter}                  & \textit{cutting}    & 83.3   & 100.0      & -- \\
\rowcolor[HTML]{EFEFEF} 
\cellcolor[HTML]{EFEFEF}                         & \textit{transport}  & 100.0  & 0.0        & -- \\
\rowcolor[HTML]{EFEFEF} 
\cellcolor[HTML]{EFEFEF}                         & \textit{handover}   & 16.7   & 0.0        & -- \\
\rowcolor[HTML]{EFEFEF} 
\multirow{-3}{*}{\cellcolor[HTML]{EFEFEF}screw}  & \textit{screwing}   & 100.0  & 0.0        & -- \\
                                                 & \textit{transport}  & 100.0  & 66.7       & -- \\
\multirow{-2}{*}{cable}                          & \textit{connecting} & 83.3   & 66.7       & -- \\
\rowcolor[HTML]{EFEFEF} 
\cellcolor[HTML]{EFEFEF}                         & \textit{transport}  & 83.3   & 0.0        & -- \\
\rowcolor[HTML]{EFEFEF} 
\cellcolor[HTML]{EFEFEF}                         & \textit{handover}   & 0.0    & 0.0        & -- \\
\rowcolor[HTML]{EFEFEF} 
\multirow{-3}{*}{\cellcolor[HTML]{EFEFEF}peeler} & \textit{peeling}    & 50.0   & 16.7       & -- \\ \hline
\end{tabular}%
}
\label{tab:world_togsuccess_uc_results}
\end{table}

The drop in performance for Binary-TOG between subsplits could suggest that objects in the UC subsplit are less commonly found in online datasets. In fact, some of the objects that Binary-TOG performed worse on include the \texttt{thermometer}, \texttt{paint roller} and \texttt{pizza cutter}. Additionally, Binary-TOG had worse performance when differentiating between \texttt{vitamins}, despite the inclusion of the bottle's color in the descriptions, which highlights the need for a textual recognition component to be added to the object recognition module. The results in Table \ref{tab:tdtog_overall_uc_results} also highlight that Binary-TOG had difficulty in distinguishing between different types of \texttt{screws} as it achieved an average F$_1$ score of 0.360, whereas OS-TOG achieved an F$_1$ score of 0.617. Since foundation models rely on generalized semantic embeddings learned from large-scale datasets, it is challenging to capture fine-grained details like size, shape, texture or, patterns even when provided with highly accurate descriptions, as opposed to OS-TOG where this information can be provided with a visual example of the object. This identified limitation also places pressure on users to ensure their textual descriptions are precise enough to represent the target object but also differentiate it from the latter categories or subcategories.

\subsection{Real-world Experiments}
The results in Table \ref{tab:world_overall_results} summarize the performance of each framework on new subcategories (KC-USC) and new categories (UC-USC) in real-world experiments with a robotic arm. Binary-TOG was able to identify and successfully grasp a target object for a target task in multi-object scenes with a success rate of 64.8\% for KC-USC objects and 52.7\% for UC-USC objects. On the other hand, OS-TOG achieved a success rate of 65.3\% for KC-USC objects and 70.9\% for UC-USC objects. Although the performance is relatively similar for the KC-USC, OS-TOG shows superior performance in generalizing to novel object categories overall.

Table \ref{tab:world_togsuccess_uc_results} showed that Binary-TOG struggled most in the UC-USC subsplit when recognizing the \texttt{screw}, \texttt{paint roller} and \texttt{peeler} by matching them to entire or partial segmentations of other objects, which is demonstrated in the first failure case of Figure \ref{fig:real_world_results}. This issue occurs more often in zero-shot techniques, as the model may focus on a prominent component of the object that encapsulates the description (e.g. the handles of scissors) rather than recognizing the entirety of the object, which disallows it to correctly identify the necessary affordance to complete the final task (e.g. \textit{handing over} the scissors).

It was also noted that for both frameworks, certain TOG success rate failures were attributed to not predicting any grasp candidates on the task-relevant region as depicted in the second failure case in Figure \ref{fig:real_world_results}.
Otherwise, failures in final TOG success rates are the results of the final grasp being unstable causing the object to slip, or the robot hitting a joint constraint. It was also seen that minor discrepancies in objects between scenes (e.g. minor rotation or placement differences) can generate a different final result, highlighting the importance of conducting multiple trials and innovation for solutions that could facilitate the recreation of identical scenes.
\section{Discussion\label{sec:discussion}}
This section discusses the experimental findings that support the primary contributions of this work, and presents their limitations and areas for future work. These contributions include: a novel multi-object dataset for training and benchmarking TOG solutions (\ref{subsec:c1}), a novel TOG framework leveraging both zero-shot and one-shot learning for object and task generalization (\ref{subsec:c2}), a new evaluation challenge focused on distinguishing between object subcategories (\ref{subsec:c3}) and robotic experiments showcasing the framework's performance in real-world scenarios (\ref{subsec:c4}). It also discusses the main research objective (\ref{subsec:o1}), which is to analyze the strengths of limitation of zero-shot and one-shot learning techniques in object generalization for TOG.

\subsection{Effectiveness of the TD-TOG Dataset}
\label{subsec:c1}
The results obtained by the baseline framework, Standard-TOG, showed that TD-TOG's labeled affordances, grasps, and object segmentations enabled the trained sub-models to achieve high TOG detection accuracy in novel scenes involving the same object instances (KC-KSC subsplit), and also operate on objects from different subcategories (KC-USC subsplit). This demonstrates that TD-TOG is effective for training TOG models, including their ability to generalize across previously unseen object subcategories. Furthermore, the comparable performance of Binary-TOG and OS-TOG achieved on TD-TOG and real-world experiments showcase that the dataset can be used as a reliable benchmark for estimating the system's expected real-world performance within a controlled setting. This significantly reduces the time and resources required to evaluate multiple TOG solutions through real-world trials.

The TD-TOG dataset is currently limited to top-down grasping scenarios. To further reflect challenges found in the real world, future work will focus on extending the dataset with more challenging settings, including 3D scenes captured from different camera angles and scenes with varying levels of clutter.

\subsection{Evaluation of Binary-TOG}
\label{subsec:c2}
Binary-TOG demonstrated competitive TOG performances across the TD-TOG dataset, especially in distinguishing between different object categories (as shown in the KC-KSC, KC-USC, UC-USC subsplits), as well as between object subcategories (KC, UC subsplits). Moreover, its performance in real-world experiments reinforces the applicability of the framework beyond datasets, validating its relevance for real-world TOG scenarios. Although Binary-TOG performed slightly worse than OS-TOG, it achieved similar inference times and requires less annotation effort by relying on textual object descriptions instead of segmentation references from the knowledge base. By adopting a modular approach for Binary-TOG, in contrast to an end-to-end framework commonly used in TOG literature, a clear identification of the performance bottlenecks was made, which were the zero-shot object recognition and one-shot affordance recognition components.

The object recognition module struggled with uncommon objects and subtle variations between subcategories, as seen in its difficulty distinguishing between different types of \texttt{screws}. This emphasizes the burden placed on users to generate highly detailed textual descriptions, however, this challenge may diminish as online datasets grow and the capabilities of foundation models advance. The affordance recognition underperformed on small objects and fragmented segments from the object segmentation module, highlighting the need for incorporating techniques that merge partial segments and use high-resolution close-ups prior to the affordance recognition module. It should be noted that preliminary experiments explored zero-shot affordance recognition solutions, aiming to develop a fully zero-shot TOG framework that eliminates the need for labeled affordance annotations in the knowledge base, but yielded suboptimal results. However, the ongoing advancements in the field of foundation models suggest this remains a promising avenue for future work. Lastly, the overall framework is currently limited to predicting top-down grasp predictions, which restricts its ability to grasp objects in an orientation that supports a task. Hence, future work explores extending the framework to use as input 3D scenes and predict 6D grasp configurations.

\subsection{Distinguishing between Subcategories}
\label{subsec:c3}
The newly proposed evaluation challenge, featured in the TD-TOG dataset as the \textit{subcategory} split, proved to be more difficult than the standard evaluation approach that assesses a model's ability to differentiate between categories as done in \textit{category} split. This is supported by the fact that both Binary-TOG and OS-TOG had worse performance for object recognition in the \textit{subcategory} split, than in the \textit{category} split. The \textit{subcategory} challenge also better reflects real-world use cases, where users often want to grasp specific subcategories rather than general categories, something that is not scalable with standard approaches like Standard-TOG as it would require training and prediction of an exponential number of subcategory labels. 

The \textit{subcategory} split also exposed limitations for both Binary-TOG and OS-TOG, particularly in their ability to differentiate between visually similar subcategories (e.g. \texttt{toothpastes}, \texttt{vitamins}, \texttt{screws}). This highlighted the potential benefit of incorporating a textual recognition component into the object recognition module as future work, which can assist in differentiating between branded objects and validate that the correct object was chosen. Hence, this novel challenge provided valuable insights by highlighting limitations and motivating future improvements, further demonstrating its effectiveness as a tool for evaluating the object generalization capabilities of TOG systems.

\subsection{Real-World Experiments}
\label{subsec:c4}
Real-world experiments demonstrated that the experimental setup and grasp predictions generated by the frameworks were successfully executed on the robotic platform. However, certain failure cases during grasp execution were observed, such as the robot reaching joint limits or objects slipping from the grippers post-grasp. In addition to the previously mentioned need to extend to 6D grasp poses (see \nameref{subsec:c2}), which would lead to improved stability, these failures highlight the benefit of incorporating semantic object properties (e.g., weight, size, material) as part of the candidate selection. Since these attributes are available in the dataset, they could be leveraged from the knowledge base after the object recognition stage to predict more stable grasp candidates within the task-relevant regions. Furthermore, future work involves exploring alternative end-effectors, such as three-fingered grippers, which may offer improved stability and support for executing the target task.

\subsection{Comparing Zero-shot and One-shot}
\label{subsec:o1}
 Results demonstrated that zero-shot learning for object recognition possesses similar generalization capabilities to one-shot learning in encountering novel objects but achieves lower accuracy. This accuracy gap arises primarily because visual-based references currently convey richer semantic information about objects, such as size, texture, and color, compared to textual references. This was exemplified when Binary-TOG performed worse in distinguishing between \texttt{screws} as there were only subtle variations between the subcategories. It can be argued that the improved accuracy from visual-based affordances used by one-shot learning comes at the cost of the manual annotation effort required, which can be tedious and require a certain level of technical expertise, limiting its accessibility. On the other hand, generating textual prompts needed by zero-shot learning that can effectively differentiate between such subtle variations, poses a large challenge for the user and may be infeasible with the current capabilities of foundation models.

Consequently, the choice between one-shot and zero-shot learning should be guided by the specific application domain. For instance, one-shot learning models are highly suitable for industrial settings, such as manufacturing, where the precise identification of specific components is critical to ensure quality and proper assembly. Additionally, manually annotating references is more feasible in these settings given there would be existing object models and annotators would have expert knowledge. Conversely, zero-shot models are more suitable for domestic applications, such as cleaning or cooking, where the user would prefer to textually describe objects rather than label every encountered object with object masks and affordances.

\section{Conclusion\label{sec:conclusion}} 
This research introduces a novel framework called Binary-TOG for task-oriented grasping (TOG). Binary-TOG builds upon our previous framework, OS-TOG, by using a zero-shot learning model instead of a one-shot learning model for the object recognition module, allowing it to generalize to novel objects through a single textual reference. In multi-object scenarios, Binary-TOG achieved an average task-oriented grasp accuracy of 68.9\%. Binary-TOG and OS-TOG are also evaluated in real-world experiments on new object categories to showcase that the TOG predictions can be executed by a robotic arm. We also present TD-TOG, a novel dataset for training and evaluating TOG frameworks. TD-TOG differs from existing datasets as it includes real-world multi-object scenes that are hand-annotated with object segment masks, affordance masks, semantic information, and grasps. This provides high-quality annotations that are aligned with the requirements of TOG solutions for both evaluation and training, and data that is representative of real-world scenarios. Additionally, the dataset features a new challenge we propose which involves assessing the ability of TOG solutions to differentiate between object subcategories, which has been largely overlooked in literature. Experimental findings show that Binary-TOG shares the same object generalization capabilities as OS-TOG, but at a reduced accuracy. Hence, it was concluded that both one-shot and zero-shot techniques are viable solutions for object generalization in TOG and should be chosen considering this performance difference, the labeling effort required when generating references, and the application domain.

Future work involves scaling the affordance recognition and grasp prediction module to detect 6D grasp poses, as well as introducing semantic-driven grasp predictions that incorporate the weight, texture, and material of the object prior to grasp execution. In addition to this, we would like to further explore the impact of the inclusion of various semantic properties in the textual descriptors for zero-shot object recognition. Additionally, we would like to add further challenges to the dataset including cluttered and 3D scenes.

\section*{Acknowledgments}
   \noindent This research is funded by a PhD studentship awarded to Early Career Researchers (Pascal Meißner) by the School of Engineering at the University of Aberdeen, UK. We gratefully acknowledge Ian Young, our lab technician at the University of Aberdeen, for assisting with the setup of the robotic platform used in real-world experiments, including the lighting and backdrop arrangements, as well as for his support in acquiring tools used in the dataset

\printbibliography

\end{document}